%% file: paper.tex
\pdfoutput=1
\documentclass[lettersize,journal]{IEEEtran}

\usepackage{amsmath,amsfonts}
\usepackage{algorithmic}
\usepackage{array}
\usepackage[caption=false,font=footnotesize,labelfont=rm,textfont=rm]{subfig}
\usepackage{textcomp}
\usepackage{stfloats}
\usepackage{url}
\usepackage{verbatim}
\usepackage{graphicx}
\usepackage{color}
\usepackage{subfloat}

\usepackage{booktabs}
\usepackage{orcidlink}
\hypersetup{hidelinks}
\def\BibTeX{{\rm B\kern-.05em{\sc i\kern-.025em b}\kern-.08em
    T\kern-.1667em\lower.7ex\hbox{E}\kern-.125emX}}
\usepackage{balance}

\begin{document}
\title{3D-aware Image Generation and Editing with Multi-modal Conditions}
\author{Bo Li$^{\orcidlink{0000-0001-9025-8985}}$, Yi-ke Li$^{\orcidlink{0000-0003-1750-9615}}$, Zhi-fen He$^{\orcidlink{0000-0003-1969-1772}}$, Bin Liu$^{\orcidlink{0000-0002-5023-7167}}$, and Yun-Kun Lai$^{\orcidlink{0000-0002-2094-5680}}$
\thanks{Manuscript created January, 2024;This work was supported in part by the Natural Science
Foundation of China (NSFC) under Grants 62172198, 61976040, 61762064,
and 62041604, in part by the Key Project of Jiangxi Natural Science Foundation
under Grant 20224ACB202008, and in part by the Opening Project of Nanchang
Innovation Institute, Peking University. The Associate Editor coordinating the
review of this manuscript and approving it for publication was XX. (Corresponding author: Zhi-fen He.)

Bo Li, Zhi-Fen He, and Bin Liu are with the School of Mathematics
 and Information Science, Nanchang Hangkong University, Nanchang
330063, China (e-mail: libonchu@outlook.com;  nyliubin@nchu.edu.cn;
 zfhe323@163.com).

Yi-ke Li is with the School of Information Engineering, Nanchang Hangkong University, Nanchang 330063, China (e-mail: 2104081200009@stu.nchu.edu.cn).
 
Yu-Kun Lai is with the School of Computer Sciences and Informatics, Cardiff
University, CF10 3AT Cardiff, U.K. (e-mail: laiy4@cardiff.ac.uk).
}}


\maketitle

\input{0_abstract}

\section{Introduction}
\label{sec:introduction}

\input{1_intro.tex}

\section{Related Work}
\label{sec:related}
\input{2_related.tex}

\section{METHODOLOGY}
\label{sec:method}

\input{3_method.tex}

\section{Experiments}
\label{sec:experiment}

\input{4_exp.tex}

\section{Conclusions}
\label{sec:conclusions}
\input{5_conclusion.tex}

\bibliographystyle{IEEEtran}
\bibliography{paper}{}




\end{document}

%% file: 0_abstract.tex
\begin{abstract}


    3D-consistent image generation from a single 2D semantic label is an important and challenging research topic in computer graphics and computer vision. Although some related works have made great progress in this field, most of the existing methods suffer from poor disentanglement performance of shape and appearance, and lack multi-modal control. In this paper, we propose a novel end-to-end 3D-aware image generation and editing model incorporating multiple types of conditional inputs, including pure noise, text and reference image. On the one hand, we dive into the latent space of 3D Generative Adversarial Networks (GANs) and propose a novel disentanglement strategy to
    separate appearance features from shape features during the generation process. On the other hand, we propose a unified framework for flexible image generation and editing tasks with multi-modal conditions. Our method can generate diverse images with distinct noises, edit the attribute through a text description and conduct style transfer by giving a reference RGB image. 
    Extensive
    experiments demonstrate that the proposed method outperforms alternative approaches both qualitatively and quantitatively on image generation and editing.     
    
     
\end{abstract}

%% file: 1_intro.tex
\IEEEPARstart{C}{onditional} image generation from a semantic label is an important and active research topic in computer vision and computer graphics. However, the challenge arises from the inherent diversity in producing plausible results for a given semantic label. Convolutional neural network~(CNN) based methods\cite{shang2016understanding} are difficult to define a proper loss function and often result in blurry images. Benefiting from the powerful adversarial learning capacity, generative adversarial networks~(GANs)\cite{goodfellow2020generative,park2019semantic} can avoid pixel-wise loss and produce photorealistic images complying with the ground truth data distribution. Pix2pix\cite{isola2017image} is one of the most typical image-to-image translation methods with conditional adversarial networks, and has sparked a wave of enthusiasm in conditional image generation. However, pix2pix based methods only conduct learning in 2D pixel spaces, and are lack of 3D consistent for multi-view generation.  

To accomplish 3D-aware image generation, EG3D\cite{chan2021efficient} proposes to first generate the 3D neural radiance field~(NeRF) representation of the image space, and then render the images from different viewpoints. EG3D extends the concept of StyleGAN\cite{Karras2020stylegan2} from 2D to 3D contexts, allowing for the generation of 3D-consistent images through the utilization of the 3D NeRF representation. Several work~\cite{sun2022fenerf,zhang20223d} attempt to conduct 3D-consistent conditional image generation from a label map by the inversion of EG3D. First, a EG3D model is trained to predict the radiance field, color field and an extra semantic field. Then, given a semantic label map, these methods try to find the optimal latent code by inverting the semantic generation branch while freezing both radiance and color fields. Finally, the conditional generation with respect to the given semantic label will be generated through a forward pass using the optimized latent code within the EG3D model. 
Nonetheless, these approaches are not complete end-to-end models, necessitating the pretraining of a NeRF model. Furthermore, these methods suffer from the gap between the inverted latent space and the pretrained GAN space. 
Other researchers have attempted to achieve 3D-aware conditional image generation by leveraging powerful pre-trained diffusion models~\cite{rombach2022high}. These methods aim to distill a 3D NeRF representation based on a given semantic label. However, most existing methods~\cite{poole2022dreamfusion,wang2023score,seo2023let,song2022diffusion} are not end-to-end models, and require separate optimization from scratch for each prompt, which is time-consuming and computationally expensive.

Pix2pix3D~\cite{deng20233d} introduces an end-to-end 3D-aware image generation model that utilizes a semantic label as conditional input. This model is a conditional 3D GAN built upon the EG3D architecture. 
The semantic label image undergoes an initial encoding process to generate a shape latent representation. This representation constitutes the initial seven features of the style vector, crucially contributing to the generation of comprehensive global geometric features.
Simultaneously, random noise sampled from a Gaussian distribution is utilized to form the remaining features of the style vector,  facilitating the creation of detailed appearance intricacies. Despite of the superior performance on 3D-consistent image generation, pix2pix3D has shortcomings in the field of image editing. First, its editing capability is limited to relying solely on the semantic label, lacking the flexibility for versatile multi-modal interactions. Second, the disentanglement performance of pix2pix3D in terms of shape and appearance editing is not satisfactory. A simulation experiment is shown in Fig.~\ref{fig:pix2pix3d_texture}. It is evident that when utilizing the same noise latent input, which is assumed to govern the appearance in pix2pix3D, the generated results exhibit significant variations in appearance properties across different semantic label conditions within each row. 

In this paper, we propose a novel end-to-end 3D-aware image generation and editing model. First, we explore the latent space of 3D GANs and propose an innovative shape and appearance disentanglement approach, which relies on a cross-attention mechanism. Second, we present a unified framework that enables flexible image generation and editing tasks incorporating diverse multi-modal conditions. This framework can produce a diverse range of unique images in response to different noise variations, accomplish attribute editing based on textual descriptions, and carry out style transfers using reference RGB images~(Fig.~\ref{fig:text_generate}).  The main contributions can be summarized as follows:
\input{fig_text_generate}
\input{fig_pix2pix3d_texture}
\begin{itemize}
    \item We explore the latent space of 3D-aware Generative Adversarial Networks (GANs) and identify meaningful latent features that are associated with semantic labels.
    \item A novel disentanglement strategy based on cross-attention mechanism is proposed to decouple appearance from shape features, ensuring consistent synthesis of appearance using the same noise latent for various semantic masks (Fig.~\ref{fig:pix2pix3d_texture}). 
    \item A multi-modal interactive 3D-aware generation framework is proposed by incorporating various conditional inputs, including pure noises, text, and reference images.
\end{itemize}

\input{fig_framework}

%% file: fig_text_generate.tex
\begin{figure*}[!t]
\centering
\includegraphics[width=\linewidth]{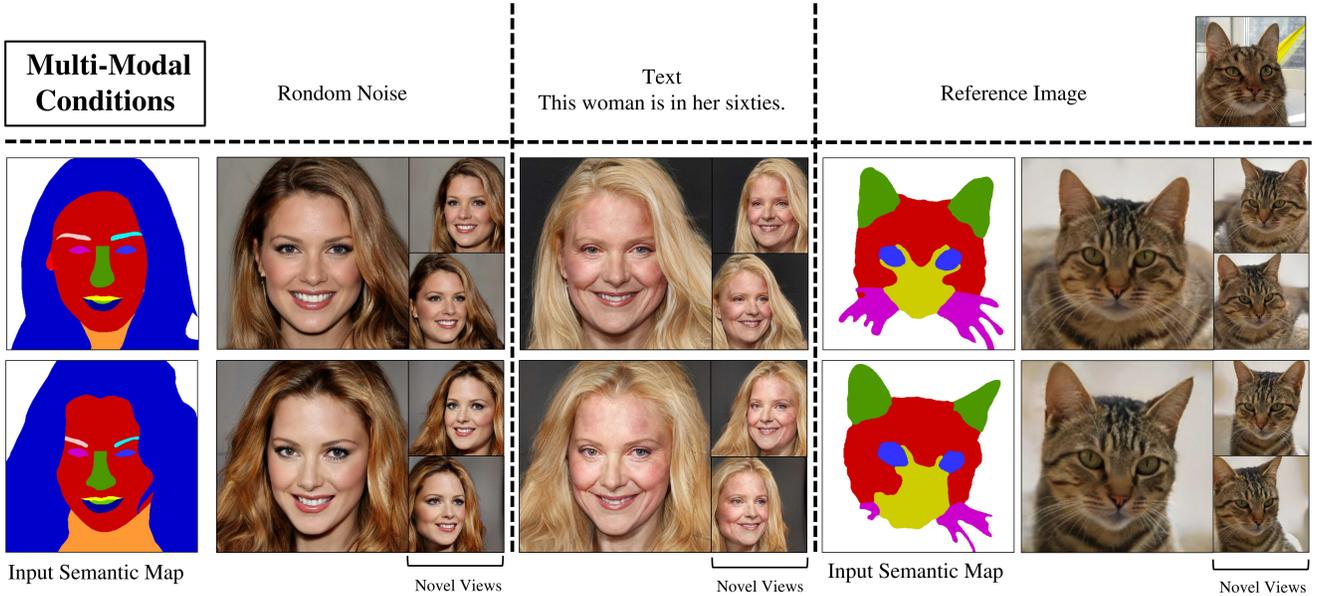}
\vspace{-2.5em}
\caption{Given a 2D semantic map, the proposed method can accomplish 3D-aware multi-view consistency generation with multi-modal conditions, such as generating diverse images with
distinct noises, editing the attribute
through a text description and conducting style transfer by giving
a reference RGB image. It is noticed that the proposed method ensures the generation of appearance consistency under a distinctive condition for various semantic maps as shown in each column. 
}
\vspace{1em}
\label{fig:text_generate}

\end{figure*}


%% file: fig_pix2pix3d_texture.tex
\begin{figure}[!t]
    \centering
    \includegraphics[width=\linewidth]{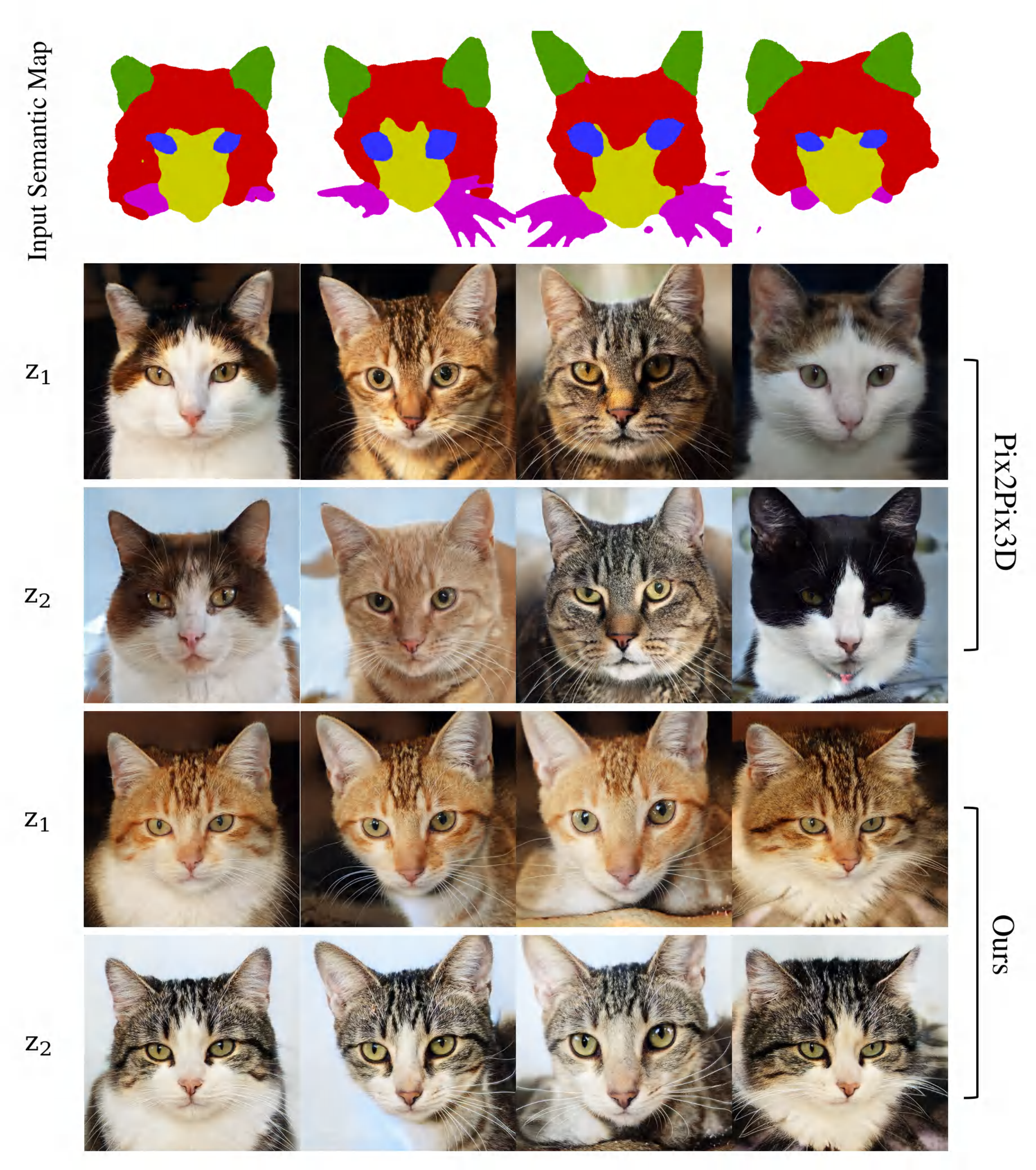}
    \vspace{-2em}
    \caption{Each row showcases the conditional image generation outcomes utilizing identical style latent variable for distinct semantic shapes. }
    
    \label{fig:pix2pix3d_texture}
\end{figure}

%% file: fig_framework.tex
\begin{figure*}[t]
    \centering
    \vspace{-2em}
    \includegraphics[width=\linewidth]{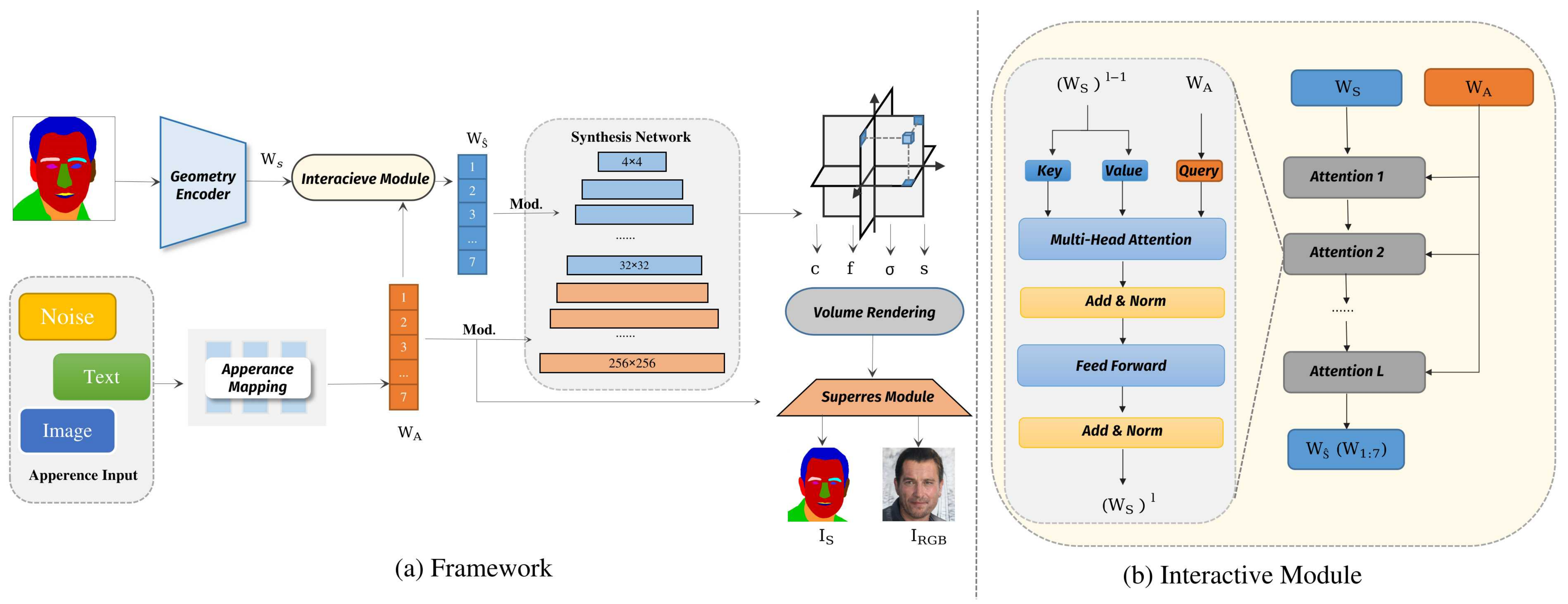}
    \vspace{-2em}
    \caption{\looseness=-1\textbf{Model overview.} (a) The framework of the proposed 3D-aware image generation model with multi-modal conditions. Given a semantic map, we can utilize a random noise, a text prompt, or a reference image as conditional inputs to accomplish 3D-aware multi-view consistency image generation. (b) The detail of the proposed interactive module.  %
    }
    \label{fig:framework2}
    \vspace{-0.5em}
\end{figure*}

%% file: 2_related.tex
\subsection{3D-aware Generative Models.}  
In contrast to 2D images, 3D objects have diverse representations, such as meshes, point clouds, occupancy, signed distance functions (SDFs), hybrids thereof and so on. The technology of generating high-quality 2D images with GAN is well-established, while it fails to capture the inherent characteristics of 3D scenes and cannot generate the underlying 3D structure of the object. Recent advancements~\cite{graf,pigan,Niemeyer2020GIRAFFE,gu2021stylenerf} have arisen through the integration of NeRF with GANs in constructing 3D-aware generators, which have leveraged different 3D representations, enabling the generation of plausible 3D structures by learning camera distribution information in parallel without explicit 3D supervision.
GRAF~\cite{graf} and $\pi$-gan~\cite{pigan} provide 3D-aware imagery and geometry generation using implicit neural rendering, but such methods achieve limited resolution.

In order to overcome this limitation, ~\cite{Niemeyer2020GIRAFFE,gu2021stylenerf,zhang2022multi,or2021stylesdf,chan2021efficient} combine 3D information, which is obtained by using volume rendering , with 2D GANs to improve the output resolution. For example, StyleNeRF~\cite{gu2021stylenerf} and MVCGAN~\cite{zhang2022multi} use a NeRF-based 3D renderer, and StyleSDF~\cite{or2021stylesdf} use a signed distance field (SDF)-based 3D renderer. To save memory, EG3D~\cite{chan2021efficient} proposes a hybrid explicit-implicit 3D representation, i.e., tri-plane 3D representation, based on 2D GANs for efficient learning of 3D distribution and reference. To reduce the need for camera parameters, a pose predictor\cite{shi2023learning} is used to simulate the pose distribution in the dataset.

\subsection{Controllable Image Generation and Manipulation.} 
To make image downstream tasks more controllable, an effective approach is to add conditional information, e.g. text \cite{xu2018attngan,zhang2017stackgan,zhang2018stackgan++}, semantic mask \cite{park2019semantic}, layout\cite{zhu2017unpaired}, etc. The StyleGAN generator ~\cite{karras2019style, Karras2020stylegan2, Karras2021} is a most widely used baseline for generating or manipulating stylized images in 2D, where the styles that control semantically meaningful properties can be learnt from random noise vectors. Some derivative work\cite{shen2020interfacegan,abdal2021styleflow,harkonen2020ganspace,shen2021closed,patashnik2021styleclip} achieve high-quality 2D conditional image generation and manipulation. However, most conditional generation and manipulation methods are mainly developed on basis of 2D GAN, without considering the idiosyncrasies of 3D objects in the latent space. Therefore, if 2D manipulation methods are directly applied in 3D space, some undesirable effects will be obtained, such as significant identity changes, degenerate facial attributes and entanglement editing.

With the development of 3DGAN\cite{chan2021efficient,graf,pigan,Niemeyer2020GIRAFFE}, some researchers devote to consider conditional input in 3D generation. Text23D \cite{cheng2023efficient} utilizes an attention-generating network operating at the word level with tri-plane features, which can synthesize fine-grained details of various sub-regions of the image. \cite{sun2022fenerf,gu2021stylenerf} use semantic map information for 3D generation, but cannot directly perform user-friendly conditional manipulation. IDE-3D\cite{sun2022ide} uses multi-branch structure to decouple geometry and texture information in different layers of the generator, but only allows editing on fixed views. Pix2Pix3D\cite{deng20233d} maps semantic graphs to the latent space of the generator which can achieve controllable semantic graph-to-image generation. Nevertheless, most of such works only consider the condition within a single modality.

\subsection{Appearance \& Shape Decoupling.} 
The research works regarding the semantically meaningful decoupling of 3D shape and appearance are relatively few. 
 The traditional 3DMM\cite{blanz2023morphable} approach uses PCA model to distinguish geometry, expression and textures in faces. Many variant algorithms of 3DMM \cite{brunton2014review,bolkart2016robust,booth20163d,cao2013facewarehouse,li2017learning,wang2017learning} are independent 3D facial texture and shape modeling methods. However, the linear characteristics of PCA makes those methods unable to effectively capture high-frequency signals. Therefore, it is difficult for 3DMMs to model the nuances of shape and texture through linear models. 
 
 With the advancement of deep learning, GAN has been used for modeling and generation of 3D shape. \cite{gecer2019ganfit,lattas2020avatarme} use GAN instead of PCA to reconstruct textures while retaining the statistical model to reconstruct shapes. \cite{gecer2020synthesizing} consider the correlation between shape and appearance, and generate high-fidelity 3D facial images by utilizing normal maps. On the other hand, \cite{gecer2018semi,sela2017unrestricted} generate photorealistic images by using style transfer GAN and 3DMM,
\cite{sun2022fenerf} divides the latent space into geometric space and texture space to achieve decoupling. But most of these methods are used for modeling and generation, making it difficult for users to input cross-modal conditions to control 3D results.

%% file: 3_method.tex
\subsection{Overview} \label{sec:overview}
In this section, we propose a novel 3D-aware conditional image generation approach from a 2D semantic label. We first dive into the latent space of 3D GANs and propose a novel disentanglement strategy to separate appearance features from geometrical shapes during the generation process. Furthermore, we propose a 3D-aware image editing algorithm with multi-modal conditions. For example, we can generate diverse images with distinct noises, edit the appearance through a text description, and conduct style transfer by giving a reference RGB image. 

The framework of the proposed method is shown in Fig.~\ref{fig:framework2}. 
Given a semantic label image $\mathbf{S}$, a geometric encoder is designed to extract the shape features $\mathbf{W}_S \in \mathbb{R}^{7\times 512}$, while the appearance features $\mathbf{W}_A \in \mathbb{R}^{7\times 512}$ can be provided by variant types of instructions, such as noises, text instructions or a reference image. Then an interactive module is proposed to estimate the appearance-aware shape features $\mathbf{W}_{\hat{S}} \in \mathbb{R}^{7\times 512}$, which is utilized to generate the shape structures with coarse appearance via modulating the first several layers of 3D GANs~\cite{chan2021efficient}. Next, the appearance code $\mathbf{W}_A$ is employed to generate the detailed texture features by adapting the rest layers of the generator. Finally, the losses including semantic similarity and appearance consistency are utilized to guide the training process.

\input{fig_stylemix}
\input{fig_stylemix_sim}

\subsection{Dissection of 3D GANs Latent Space} \label{sec:latent_style}
In this section, we dive into the latent space of 3D GANs~\cite{chan2021efficient}, and propose some meaningful discoveries of semantic control during the generation process. 
As the generator of EG3D~\cite{chan2021efficient} is very similar to the one of StyleGAN\cite{Karras2020stylegan2}, 
the style vectors within the latent space possess significant semantic control capabilities. 
Although EG3D~\cite{chan2021efficient} has undertaken certain initiatives to showcase the semantic nature inherent in the latent space, they only demonstrate a core idea that the majority of the structures are from the modulations of the first several layers, while the details of the shape and textures are from the middle-last layers. However, the detailed and quantitative analysis are not provided.

In this section, we first conduct a more detailed style mixing experiment, and then propose a quantitative metric based on semantic labels to explore the semantic latent space. Specifically, we randomly sample two latent codes $z_a$ and $z_b$, and the corresponding style codes $\mathbf{W}^{a} \in \mathbb{R}^{14\times 512}$ and $\mathbf{W}^{b} \in \mathbb{R}^{14\times 512}$ and generated images $\mathbf{I}^{a}$ and $\mathbf{I}^{b}$ as done in StyleGAN\cite{Karras2020stylegan2} can be obtained. To unveil the meaningful semantic properties embedded in each dimension of the style vector, we conduct style mixing experiments by blending the style vectors $\mathbf{W}^{a}$ and $\mathbf{W}^{b}$ at a selected intersection point $N$, $\mathbf{W}^{a\rightarrow b} = [\mathbf{W}^{a}_{1:N}, \mathbf{W}^{b}_{N+1:14}]$.
The mixing results $\mathbf{I}_{a\rightarrow b}^{N}$ with different intersections $N$ are shown in Fig.~\ref{fig:eg3d_style_mixing}. We can find that $N=7$ is a good candidate. The first seven layers mainly  focus on the prominent geometrical attributes, such as face shape, hair length, glasses and so on. While the rest layers control the appearance features, such as hair color, skin color, lighting and so on.

To validate the meaningful phenomenon, we conduct quantitative experiments to analyze the latent space in detail. As this paper focus on 3D-aware image generation with a semantic label as condition, the meaningful latent space relevant to semantic labels is explored. Let $SIM_A$ denotes the similarity between the semantic maps of the style mixture outcomes $\mathbf{I}_{a\rightarrow b}^{N}$ and those of $\mathbf{I}_{a}$, and  $SIM_B$ represents the similarity between the semantic maps of $\mathbf{I}_{a\rightarrow b}^{N}$ and those of $\mathbf{I}_{b}$, $N$ is the intersection value.  $SIM_A$ and  $SIM_B$ can be defined as follows:

\begin{eqnarray}
    SIM_A=PA[f_{S}(\mathbf{I}^{a}),f_{S}(G(\mathbf{W}^{a}_{1:N}, \mathbf{W}^{b}_{N+1:14}))]\\
    SIM_B=PA[f_{S}(\mathbf{I}^{b}),f_{S}(G(\mathbf{W}^{a}_{1:N}, \mathbf{W}^{b}_{N+1:14}))] 
\end{eqnarray}
where Pixel Accuracy (PA) denotes the percent of pixels that are accurately classified within the image, which is computed by accounting the ratio between the amount of accurately classified pixels and the total number of pixels in the image. $f_{S}(\cdot)$~\cite{yu2018bisenet,amir2021deep} is a pre-trained semantic map extractor, $G(\cdot)$ is the style mixing image generator.

Furthermore, to assess the variance between adjacent intersections, the semantic shape variability $SIM_N$ within the style mixing outcomes across neighboring layers $N-1$ and $N$ is defined as follows:
    
\begin{eqnarray}
    SIM_N=PA [f_{S}(G(\mathbf{W}^{a}_{1:N-1}, \mathbf{W}^{b}_{N:14})), \nonumber\\
    f_{S}(G(\mathbf{W}^{a}_{1:N}, \mathbf{W}^{b}_{N+1:14}))]
\end{eqnarray}


The average quantitative metrics results on CelebaMask and AFHQ datasets are shown in  Fig.~\ref{fig:eg3d_style_mixing_curve}. 
Given an image generated by the latent $\mathbf{W}^b$, the progression of style mixing showcases a shift from the semantic (shape) features of $\mathbf{I}^b$ towards those of $\mathbf{I}^a$.
It can be clearly observed that this transition tends to closely resemble $\mathbf{I}^a$ as the process advances, particularly evident from $N=7$ onwards.
Furthermore, the curve of metric $SIM_N$ also indicates that the variance between adjacent intersections tends to be stable starting from $N=7$. 
The above quantitative results demonstrate that $N=7$ is a good candidate for decoupling of semantic shape space and texture space. 

However, it is noteworthy that the initial 7 layers are not exclusively dedicated to shape changes, they also encompass texture-related variations. This is our second meaningful discovery. One of the most related work, pix2pix3D\cite{deng20233d}, propose to generate the shape structure by exclusively utilizing the semantic shape feature to form the initial 7 layers of the style vector, and employ a Gaussian latent representation for the last 7 layers of the style vector to control the generation of appearance characteristics.
It seems to decouple the shape and appearance well from the model designing, i.e., the same Gaussian noise should generate the same appearance for different semantic conditions. However, based on our second discovery, we find that not only the basic shape but some random coarse appearance features will be generated by the first 7 layers of the style vector, and the modulations by the last several layers with the Gaussian noise can only refine the coarse texture rather than completely dictating the appearance generation. Consequently, given a fixed Gaussian noise, the appearance of the generated results based on different semantic maps are distinct, such as shown in Fig.~\ref{fig:pix2pix3d_texture}. To achieve disentangled generation of shape and appearance, a novel strategy will be introduced in the next section.       


\subsection{Disentangled Generation of Shape and Appearance}\label{sec:adapter}
Building on the significant findings from the previous section, pix2pix3D\cite{deng20233d}'s inadequate performance on disentangled generation can be attributed to the exclusive reliance of the initial 7 layers' coarse generation on modulation by the shape feature $\mathbf{W}_{S}$. However, this method generates arbitrary appearance due to the lack of interaction with the appearance code $\mathbf{W}_{A}$.
In this section, we introduce an innovative strategy aiming at ensuring the disentangled generation of shape and appearance. As the initial seven layers' generation includes both geometric shape and coarse appearance, we propose to learn an appearance-aware shape feature $\mathbf{W}_{\hat{S}}$ via a cross-attention based interaction module. The proposed method can achieve disentangled generation and editing. For a fixed appearance latent $\mathbf{W}_A$, given different semantic label images $\{\mathbf{S}_i\}$, the proposed interactive module will predict similar appearance features for each $\mathbf{S}_i$ to the same attribute $\mathbf{W}_A$, rather than generating a random coarse appearance like in pix2pix3D\cite{deng20233d}. 

Furthermore, pix2pix3D can only conduct image generation and editing with a semantic label but not deal with diverse editing tasks. In this paper, we propose a multi-modal interactive 3D-aware generation framework incorporating various conditional inputs, including pure noises, text, and reference images. 

Next, we introduce the detailed descriptions of each module.

\textbf{Feature Encoder.} 
A semantic map encoder $\mathbf{E}_s$ is designed to learn the shape embedding feature $\mathbf{W}_{S}$ from a given semantic label image $\mathbf{S}$. $\mathbf{E}_s$ is a simple multi-scale convolution neural network, covering a range from $512 \times 512$ to a 1D global feature.

For the task of conditional generation, we use the style-based generators\cite{karras2019style,chan2021efficient} to map Gaussian noise to the $w$-space $\mathbf{W}_N$. In order to inject multi-modal conditions, we design text encoder $\mathbf{E}_T (\cdot)$ and image encoder $\mathbf{E}_I (\cdot)$ respectively to get the corresponding embedding $\mathbf{W}_T$ and $\mathbf{W}_I$. 
$\mathbf{E}_T$ adopts a two-layer LSTM network with cell size of 1024 and two fully-connected layers following the LSTM, while $\mathbf{E}_I$ use same network as $\mathbf{E}_s$.
We concatenate features $\mathbf{W}_N$ and $\mathbf{W}_T$ or $\mathbf{W}_I$ through an additional adapter network to obtain a conditional appearance latent $\mathbf{W}_A$ with corresponding text or reference image information.

\textbf{Interaction Module with Cross Attention}
As discussed in Sec.~\ref{sec:latent_style}, the insufficient disentanglement performance primarily stems from the sole dependence of the initial 7 layers' rough generation on modulation by the shape feature $\mathbf{W}_{S}$. Therefore, a novel interaction module (Fig.~\ref{fig:framework} (b)) based on cross attention is proposed, which is used to predict the appearance-aware shape features to guide the preliminary generation with precise shape and coarse appearance.

To find the relevance between the appearance feature $\mathbf{W}_A$ and the shape feature $\mathbf{W}_S$, we set the embedding of the appearance feature $\mathbf{W}_A$ as the query~(Q), and utilize the embedding of shape feature  $\mathbf{W}_S$ as key~(K) and value~(V). The interaction layer in each attention module can be represented as follows,  
\begin{equation}
    \mathbf{Q}=\mathbf{W}_A \mathbf{W}^Q, \mathbf{K}=\mathbf{W}_S \mathbf{W}^K, \mathbf{V}=\mathbf{W}_S \mathbf{W}^V,
\end{equation}
\begin{equation}
    \mathbf{W}_{S}^{l+1}=softmax(\frac{\mathbf{Q} \mathbf{K}^T}{\sqrt{d_k}}) \mathbf{V}+(\mathbf{W}_S)^l
\end{equation}
where $\mathbf{W}_Q$, $\mathbf{W}_K$ and $\mathbf{W}_V$ are the linear projection matrices, and $d_k$ is the common dimension of the latent codes in the projection, which is set to 128. 
Through the proposed interaction module, we can get the appearance-aware shape feature $\mathbf{W}_{\hat{S}}$, served as the initial 7 layers
of the style vector to produce the preliminary generation result
with precise shape and coarse appearance.
Meanwhile, the appearance feature $\mathbf{W}_A$ is used to modulate the rest layers of the generator to refine the appearance. Similar to pix2pix3D, the generator will output the tri-plane features, which are then encoded to both the semantic feature $s$ and radiance feature $(c, f, \sigma)$, where $c$ is the RGB color of the sampling point, $\sigma$ is a density value, and f is a 64-dimensional feature used for super-resolution. Finally, given a viewpoint $d$, a semantic map $\mathbf{I}_S$ and a RGB image $\mathbf{I}$ will be rendered respectively.

Equipped with the proposed interaction module, we can generate images with consistency appearance with the same noise for distinct semantic label maps~(Fig.~\ref{fig:pix2pix3d_texture}). While the existing methods, e.g., pix2pix3D, will generate distinct appearances for different semantic masks with a same appearance feature~(Fig.~\ref{fig:pix2pix3d_texture}). Moreover, the proposed method can accomplish image editing with multiple modals of conditions, such as text instructions and reference images.

\input{fig_compare}

\subsection{Training Objectives}\label{sec:loss}

The training dataset is composed of paired data ($\mathbf{I}$, $\mathbf{S}$, $d$), where $\mathbf{I}$ and $\mathbf{S}$ are the aligned RGB image and the corresponding semantic label image respectively, and $d$ is the ground truth camera pose. For text-conditioned image generation, an additional aligned text instructions is required.

Learning a 3D-aware conditional image generation and editing is a challenging task.
We systematically devise a constellation of learning objectives, including  adversarial loss, semantic reconstruction and cross-view consistency losses.

\textbf{Adversarial Loss.}
To ensure the generated images maintain a realistic appearance from any arbitrary angle. 
The adversarial losses are defined on both generated RGB image $\mathbf{I_{RGB}}$ and the semantic label $\mathbf{I_s}$. Therefore, the model has two discriminators $D$ and $D_s$ for RGB images and label maps, respectively. They concat high-resolution output and low-resolution output together when calculating the adversarial loss.

The discriminator $D$ takes both real and generated fake images as input, and $D_s$ concatenates color images $I_{RGB}$ and the label maps $I_s$ as input to facilitate pixel alignment of generated images and the corresponding semantic maps. The adversarial loss can be represented as follows:

\begin{eqnarray}
    \mathcal L_{adv} =\mathcal L_D(I_{RGB},I_{RGB}^+)+\nonumber\\
    \lambda_{D_s} \mathcal L_{D_s}(I_{RGB},I_{RGB}^+,I_s,I_s^+)
\end{eqnarray}

\textbf{Cross-view Consistency Loss.} In order to ensure the consistency of the 3D shape generated by the semantic map from different views, we use a multi-view consistent loss for the semantic map as done in pix2pix3D~\cite{deng20233d},
\begin{equation}
    \mathcal L_{cvc} = \lambda_c \mathcal L_1(G(R(G(\mathbf{S},p),p'),p),\mathbf{S})
\end{equation}
where $\mathbf{S}$ is the semantic map, $p$ and $p'$ are two distinct viewpoints, $R(\cdot)$ is the render function, $G(\cdot)$ is the tri-plane generator, $\mathcal L_1(\cdot)$ is the $L_1$ distance.

\textbf{Semantic Reconstruction Loss}. The adversarial loss is not enough to guide the semantic alignment between multi-modal conditions and generated results. Hence, we propose a balanced cross-entropy loss $\mathcal{L}_s$ to regulate both the input semantic map $\mathbf{S}$ and the reconstruction outcomes $\mathbf{I}_s$ and $\mathbf{I}_{s}^{+}$ at varying resolutions.
Meanwhile, we calculate the LPIPS loss for the rendered RGB Images at various resolutions, represented as $\mathbf{I}_{RGB}$ and $\mathbf{I}_{RGB}^{+}$,
\begin{equation}
    \mathcal L_{recon}= \mathcal L_{lpips}(\mathbf{I},\{\mathbf{I}_{RGB},\mathbf{I}_{RGB}^{+}\})+ \mathcal L_s(\mathbf{S},\{\mathbf{I}_s,\mathbf{I}_s^+\})
\end{equation}

\textbf{Appearance Contrastive Loss}. 
To enhance the separation of geometry and texture, we employ a contrastive loss to
bring the style features of images with identical input appearance features closer, while ensuring that styles with different appearance features are kept separate. In this context, we leverage the Gram matrix~\cite{gatys2016image} of deep features to constrain the texture features of the generated image,
\begin{equation}
    \mathcal L_{ac}= \lambda_t \frac{\mathcal L_{Gram}(G(S_i,z),G(S_j,z))}{\mathcal L_{Gram}(G(S,z_i),G(S,z_j))} 
\end{equation}

\textbf{Multi-modal Consistency Loss.} 
The proposed method cannot only accomplish diverse image generation with a given semantic map $\mathbf{S}$ and various latent noise $z$, but can conduct multi-modal interactive generation and editing, including text instructions and reference images.

For text-conditioned generation tasks, we leverage a pre-trained fine-grained image classification network $C$\cite{jiang2021talk} to help establish an additional text-semantic consistency loss.
Within the CelebAMask-HQ dataset, every image is paired with a text description $T$ and an associated attribute label $L$, indicating the strength of the trait being described.   
Given a text description $T$ and its fine-grained label $L$, the semantic consistency loss is defined as:
\begin{equation}
    \mathcal L_{modal}= \mathcal L_{1}(L,C(G(\mathbf{S},T,z)))
\end{equation}
where $C(\cdot)$ is a pre-trained  model to predict the fine-grained attribute label $L$.

For the task of reference-guided generation, the cycle-consistency loss is proposed to constrain the generated images with similar appearance for the given reference image $\mathbf{I}_t$. First, the latent appearance feature of the generated image $G(\mathbf{S}, \mathbf{I}_t)$ should be consistent to the one of the reference image $\mathbf{I}_t$. Second, given the ground truth semantic maps $\mathbf{S}_t$ of the reference image $\mathbf{I}_t$, the synthesized image with the guidance of the generated image $G(\mathbf{S}, \mathbf{I}_t)$ should reconstruct the ground truth reference image $\mathbf{I}_t$ as good as possible. Therefore, the cycle-consistency loss is defined as follows:

\begin{eqnarray}
      \mathcal L_{modal} &=& \mathcal L_{1}(E(\mathbf{I}_t),E(G(\mathbf{S},\mathbf{I}_t))) \nonumber \\&+& \lambda_m \mathcal L_{1}(\mathbf{I}_t,G(\mathbf{S}_t,G(\mathbf{S},\mathbf{I}_t)))
\end{eqnarray}

\textbf{Overall Training Objective.} The total objective is the weighted sum of the above five losses,
\begin{equation}
    \mathcal L_{total}=\mathcal L_{adv}+ \mathcal L_{recon}+ \mathcal L_{cvc}+ \mathcal L_{ac}+ \mathcal L_{modal}
\end{equation}

\input{fig_compare_cat}

%% file: fig_stylemix.tex
\begin{figure}[!t]
        \includegraphics[width=\linewidth]{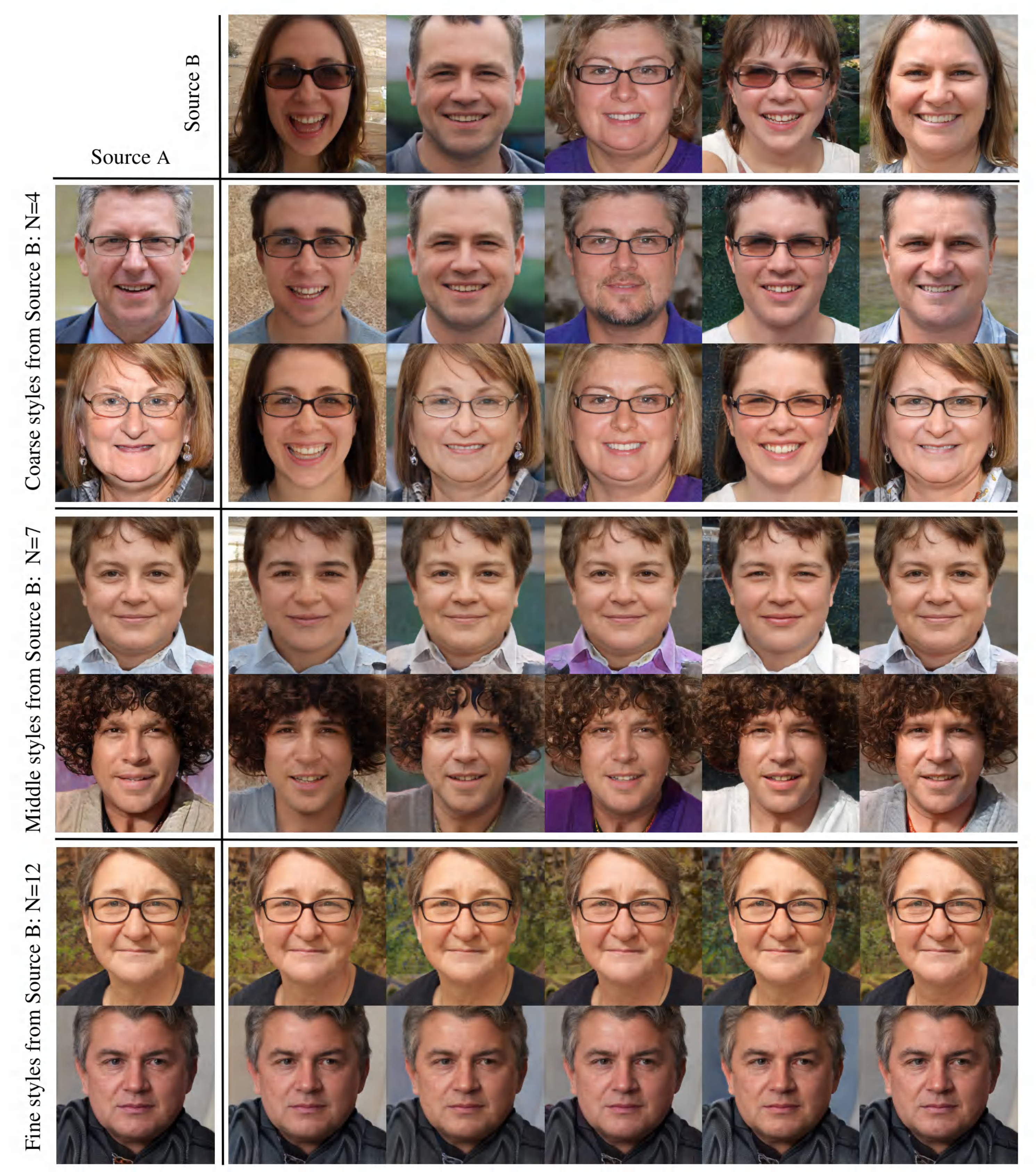}
    \vspace{-2em}
    \caption{Style mixing results in the latent space of EG3D\cite{chan2021efficient}.  Given a source image A (on the first column) and a target image B (on the top row), the mixed images are synthesized by blending the style vectors of A and B at a selected intersection point N.  } 
   
    
    \vspace{-1.5em}
    \label{fig:eg3d_style_mixing}
\end{figure}

%% file: fig_stylemix_sim.tex
\begin{figure}[!t]
        \centering
        \subfloat[Semantic Structure Similarity on CelebaMask dataset.]{\includegraphics[width=\linewidth]{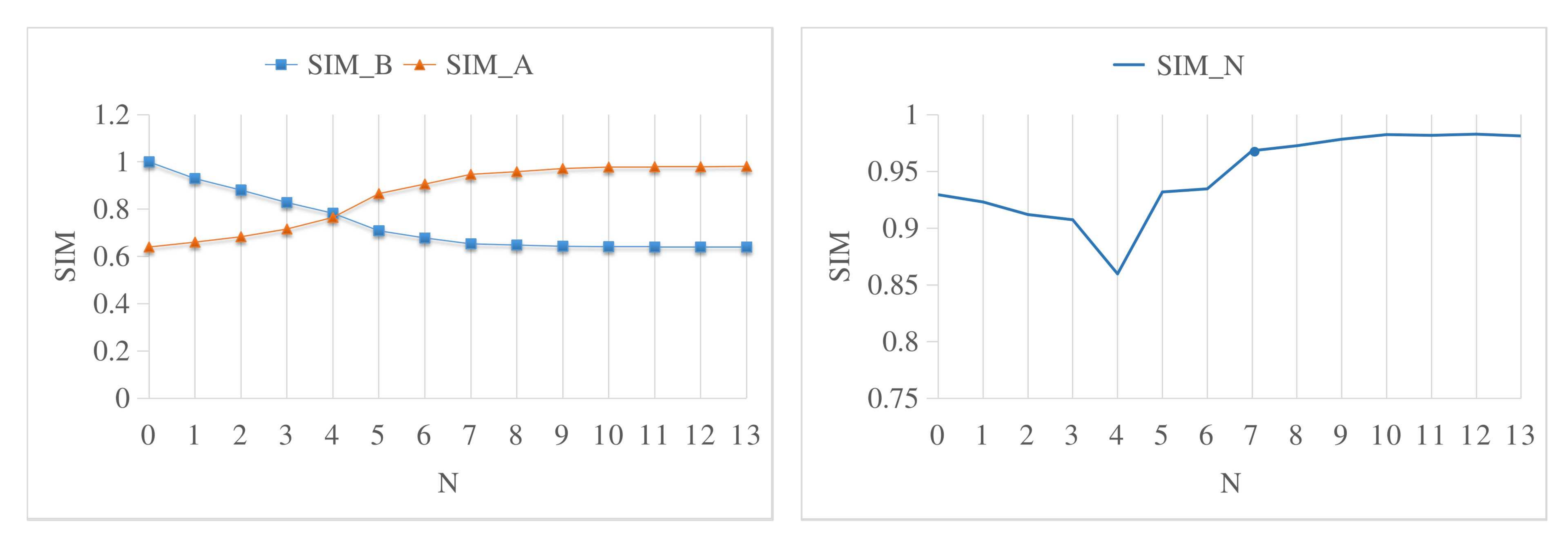}}%

        \subfloat[Semantic Structure Similarity on CelebaMask dataset.]{\includegraphics[width=\linewidth]{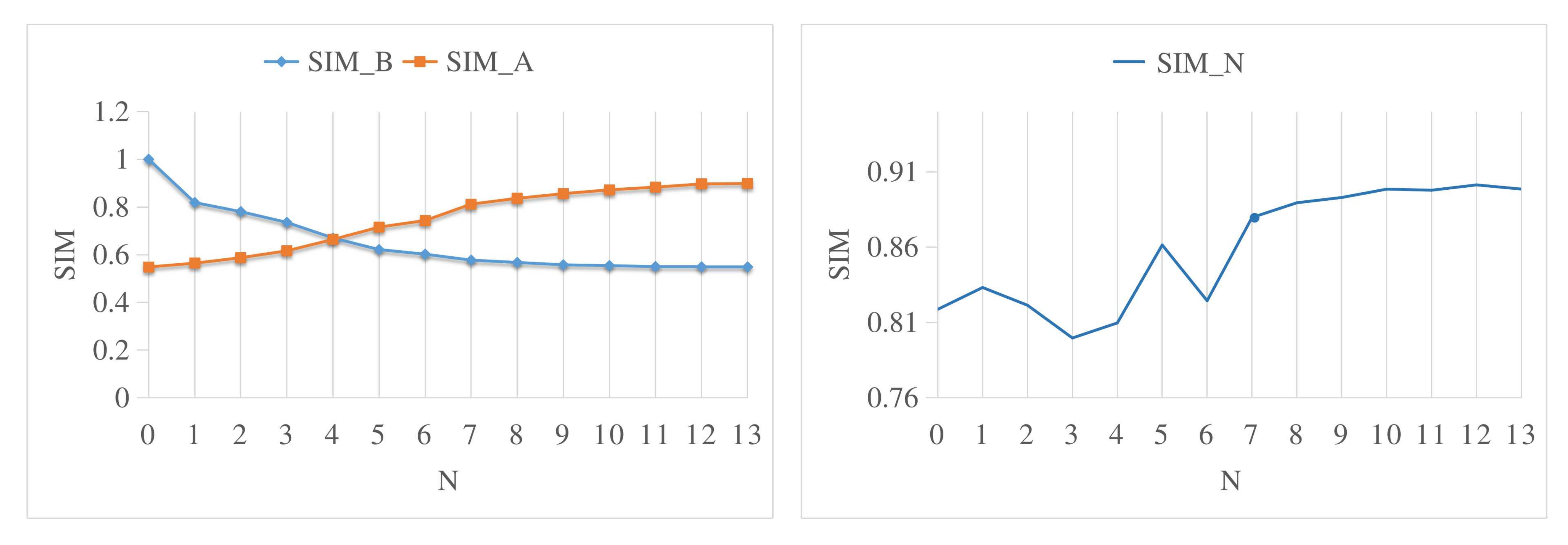}}
        
    \caption{Quantitative evaluations of semantic structure similarity of style mixing in the latent space of EG3D\cite{chan2021efficient}.} 

    \vspace{-1.5em}
    \label{fig:eg3d_style_mixing_curve}
\end{figure}

%% file: fig_compare.tex
\begin{figure*}[t!]
    \centering
    \vspace{-1em}
    \includegraphics[width=0.75\linewidth]{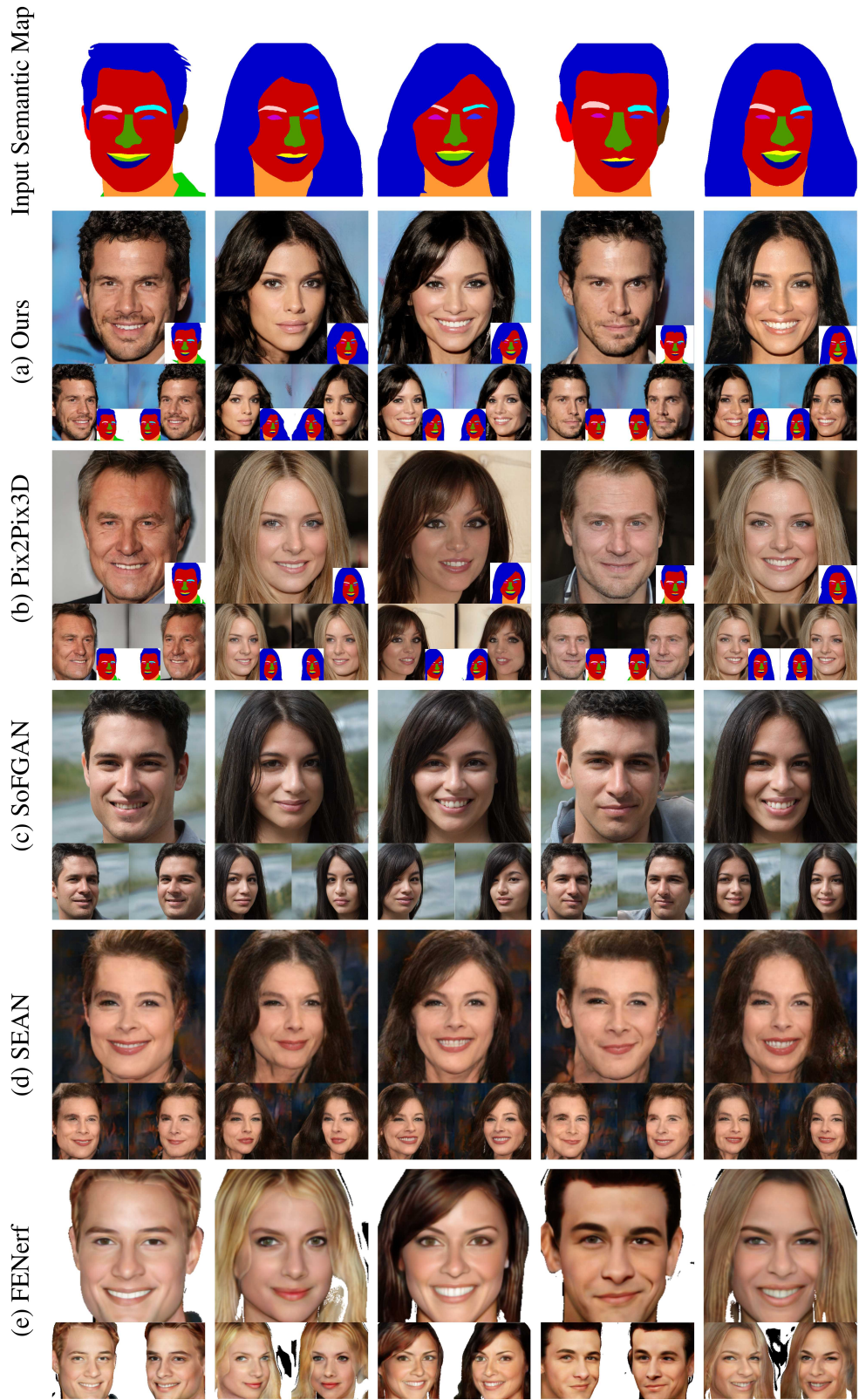}
    \vspace{-2em}
    \caption{Visual inspection of conditional generation results by different methods on CelebAMask dataset.  %
   } 
    \label{fig:compare}
    \vspace{-1em}
\end{figure*}

%% file: fig_compare_cat.tex
\begin{figure}[t!]
    \centering
    \includegraphics[width=\linewidth]{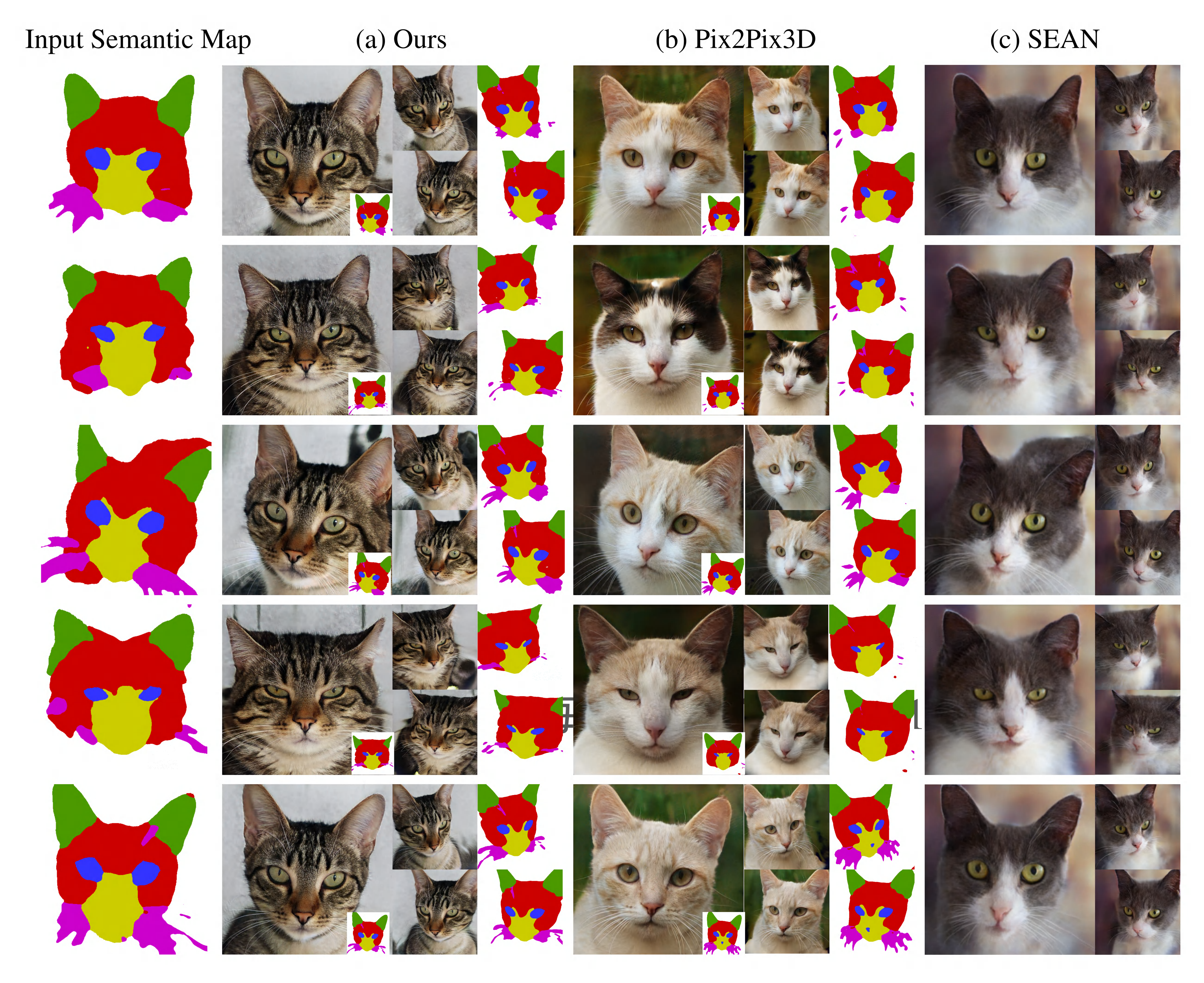}
    \vspace{-2em}
    \caption{Visual inspection of conditional generation results by different methods on AFHQ-Cat dataset.  %
    }
    \label{fig:compare_cat}
    \vspace{-1em}
\end{figure}

%% file: 4_exp.tex

We firstly introduce the datasets and baseline methods in Sec.~\ref{sec:training_ditails}, then the quantitative evaluation and ablation experiments are conducted to evaluate the effectiveness of the interactive module in Sec.~\ref{sec:Qualitative Evaluation} and ~\ref{sec:Ablation Study}. Finally, we demonstrate the method’s applications in multi-modal condition generation in Sec.~\ref{sec:Applications}.

\subsection{Implement details} \label{sec:training_ditails}
\textbf{Datasets.} CelebAMask-HQ~\cite{lee2020maskgan} contains 30k face images at 1024×1024 resolution, and each image has a corresponding facial semantic segmentation mask which contains 19 categories, including eyes, hairs, ears, nose, lips, etc. Following~\cite{deng20233d}, we predict the pose associated with each image by HopeNet\cite{doosti2020hope}. CelebADialog\cite{jiang2021talk} provides fine-grained attribute labels and corresponding captions for each image. Each semantic attribute is divided into 6 levels according to the degree of the feature. We select two categories: age and beard for text condition generation experiments.
AFHQ-CAT~\cite{choi2020starganv2} comprises 5,065 images of cats at 512×512 resolution. We segment each cat image into six semantic categories, including ears, eyes, beard, nose, face shape and background by\cite{amir2021deep}. 

\input{fig_ablation}
\textbf{Training Details.} 
We initialize our model with part of pretrained weights from EG3D\cite{chan2021efficient}. To further improve the stability of the fine-tuning process, a two-stage strategy is adopted. In the first stage, the model undergoes training solely on GAN loss and semantic label reconstruction loss, progressively adjusting the conditional feature encoder and interaction module to suit the training dynamics. During the second stage, we incorporate text-semantic consistency loss and cycle-consistency loss to align text instruction or reference image with the generated 3D-aware images. During the first stage of training, 500k images are utilized, while the second stage involves 1,000k images. The batch size is set as 8 in all experiments, and the model is trained with 6 hours using 8 Tesla V100 GPUs. The hyperparameters are set as $\lambda_{D_s}=0.1$, $\lambda_{c}=0.0001$ and $\lambda_{m}=\lambda_t=0.1$.

\input{tab_seg2face}
\input{tab_seg2cat}

\textbf{Baselines.}
In this paper, we compare the proposed method with four state-of-the-art 2D and 3D-aware conditional generation approaches based on semantic maps: pix2pix3D\cite{deng20233d}, SoFGAN\cite{chen2022sofgan}, SEAN\cite{zhu2020sean} and FENeRF\cite{sun2022fenerf}. Pix2pix3D is the most related work to the proposed method, which is an end-to-end 3D-aware conditional image generation model. SoFGAN uses a 3D semantic map generator to generate semantic maps of different poses to generate 3D-aware face information. SEAN is a state-of-the-art 2D conditonal GAN model to generate high-quality images with a semantic map as condition. FENeRF conducts 3D-consistent conditional image generation from a label map by the inversion of a 3D GAN. 


\subsection{Quantitative and Qualitative Evaluation} \label{sec:Qualitative Evaluation}
In this section, we evaluate the proposed method from four aspects, including image quality, semantic alignment, multi-view consistency and appearance consistency. 

Firstly, to assess the quality of the generated images, we randomly generate 5k images and compute the Fréchet Inception Distance(FID)\cite{heusel2017gans} and Kernel Inception Distance (KID)\cite{binkowski2018demystifying} between the generated and ground truth images. In addition, the LPIPS between generated pairs using identical conditional semantic map but different noise inputs, is calculated for diversity evaluation.

Secondly, to evaluate the alignment between the generated images and the given semantic label, we randomly selected 1k segmentation maps as conditional input from the test dataset and compute the mean Intersection-over-Union (mIoU) and pixel accuracy (acc) between the input semantic map and the reconstructed segmentation maps.

Thirdly, to assess the multi-view consistency of the free-viewpoint images generated by various methods, which is a crucial aspect of 3D-aware generation models' performance, the FVV Identity~\cite{deng20233d} between generated free-viewpoint images with the same semantic map $\mathbf{S}$ and the appearance code $z$ with the face recognition Network Facenet\cite{schroff2015facenet} is calculated. In addition, to ensure identity consistency, we also assess appearance consistency by measuring the distance between the Gram features of images generated using the same appearance code $z$ but two different semantic maps.


The quantitative experimental results on both datasets are listed in  Table.~\ref{table:face_comparision} and Table.~\ref{table:cat_comparision}. It is obviously that the proposed method achieves the best performance on image quality. Moreover, our proposed method achieves superior pixelwise alignment and comparable performance in mIoU compared to pix2pix3D on semantic alignment. Despite similar image quality performance, the proposed method outperforms in both multi-view consistency and texture consistency. This advantage stems from the proposed disentangling strategy. With a fixed appearance code $z$ and varying semantic maps as conditions, the proposed method generates images with diverse shapes but consistent appearances, as opposed to pix2pix3D, which generates random textures. 

\input{fig_text_age}

In addition to quantitative comparisons, the visual inspection results of different methods are shown in 
Fig.~\ref{fig:compare} and Fig.~\ref{fig:compare_cat}. It is noticed that the synthesized results in each column are generated by a fixed latent vector $z$. It is obvious to find that pix2pix3D~\cite{deng20233d} can generate realistic and consistency multi-view images. Nevertheless, the generated outcomes with the identical latent input lack consistency in appearance, e.g., the color of hair and skin, leading to a lack of control over the results.
Although SoFGAN achieves the best accuracy of semantic alignment on the mIoU metric, it suffers from the multi-view consistency, notably in areas like the beard and teeth regions. Compared with the existing methods, our proposed approach excels in generating images that maintain both multi-view consistency and appearance uniformity across diverse semantic label conditions using a single latent input. Therefore, it has high adaptability and flexible generative process.


\input{fig_image_cat}
\input{fig_stylemix_compare}
\input{fig_interp}
\subsection{Ablation Study} 
\label{sec:Ablation Study}
In this section, we conduct an ablation study to evaluate the influence of the interactive module proposed in Sec.~\ref{sec:adapter}, which is crucial to the disentangled shape and appearance generation in our architecture.  In this experiment, we remove the interactive module, and utilize the latent shape feature $\mathbf{W}_S$ to modulate the first 7 layers of the generator, while the rest layers are adapted by the appearance code $z$ as done in pix2pix3D. The experimental results are shown in Fig.~\ref{fig:ablation}. The generative results in each column share an identical appearance code. In the absence of the interactive module, the initial 7 layers yield coarse generative results, including shape structure and random coarse appearances, like fur color patterns. Consequently, even with the same appearance code, the model generates images with distinct textures, evident in each column. Conversely, the model incorporating the proposed interactive module can improve texture consistency in its generated images, which validate the performance of the proposed module.

\subsection{Applications} 
\label{sec:Applications}

\textbf{Multi-modal Conditional Generation.} 
Compared with pix2pix3D, the proposed method cannot only conduct 3D-aware image generation with a semantic map as condition, but can accomplish image editing with multi-modal conditions, such as text and reference image.

As shown in Fig.~\ref{fig:text_age}, given a semantic map, we can edit the attribute `age' or `beard' using a piece of text instruction. It is obvious to find that the proposed method can achieve 3D consistency editing results.

The proposed method can also transfer the texture of a reference RGB image to the generated results. Some results are shown in Fig.~\ref{fig:image_cat}. 
It's evident that the proposed method excels in generating consistent textures identical to the reference image across different semantic maps. For pix2pix3D, we conduct a similar experiment by substituting the appearance code with the encoded appearance feature from the reference image, and the results are shown in the last column of Fig.~\ref{fig:image_cat}. It can be found that pix2pix3D will generate results with random textures. This phenomenon further validates the motivation of the proposed model. 




\textbf{Latent Space Interpolation} 
In this section, we conduct two interpolation experiments to validate the latent space consistency. 

In the first experiment, we conduct style mixing experiments by blending the style vectors $\mathbf{W}^{a}$ and $\mathbf{W}^{b}$ at a selected intersection point $N$, $\mathbf{W}^{a\rightarrow b} = [\mathbf{W}^{a}_{1:N}, \mathbf{W}^{b}_{N+1:14}]$. The style mixing results of both pix2pix3D and the proposed method are shown in Fig.~\ref{fig:eg3d_style_mixing_compare}.
An effective disentanglement approach should yield a clear separation between shape and appearance, as done by the proposed method (Fig.~\ref{fig:eg3d_style_mixing_compare} (b)), only the shape structures exhibit variations from `a' to `b' until N=7 in the style mixing results, while changes in appearance occur within the last 7 layers.
However, the outcomes from pix2pix3D (Fig.~\ref{fig:eg3d_style_mixing_compare} (a)) demonstrate the entangling of both shape and appearance within the initial 7 layers.

In the second experiment, we conduct latent interpolation along shape feature $\mathbf{W}_S$ and appearance feature $\mathbf{W}_A$. Each row showcases results generated using the identical shape feature alongside interpolated appearance features. Conversely, each column represents outcomes generated with the same appearance code and interpolated shape latent.
It is obvious to find that the appearance keep consistency while the shape varies from `a' to `b' distinctly within each column of the results generated by the proposed method (Fig.~\ref{fig:interp}(a)). However, pix2pix3D encounters challenges in effectively disentangling shape and appearance, notably showcased in the first column of Fig.~\ref{fig:interp}(b), where the appearance significantly changes during the interpolation of shape latent. This experiment underscores the limitation of pix2pix3D's generative strategy in adequately disentangling shape and appearance, leading to a lack of flexibility and controllability in generating and editing.



\textbf{Single-view 3D Reconstruction.} Fig.~\ref{fig:3d} demonstrates the application of our model in single-view 3D reconstruction. The reconstructed 3D model shows that our model can achieve high-quality geometry recovery while preserving semantic information. The extraction of 3D information is executed by marching cube algorithm.

\input{fig_3d}

%% file: fig_ablation.tex
\begin{figure}[t!]
\centering
\includegraphics[width=\linewidth]{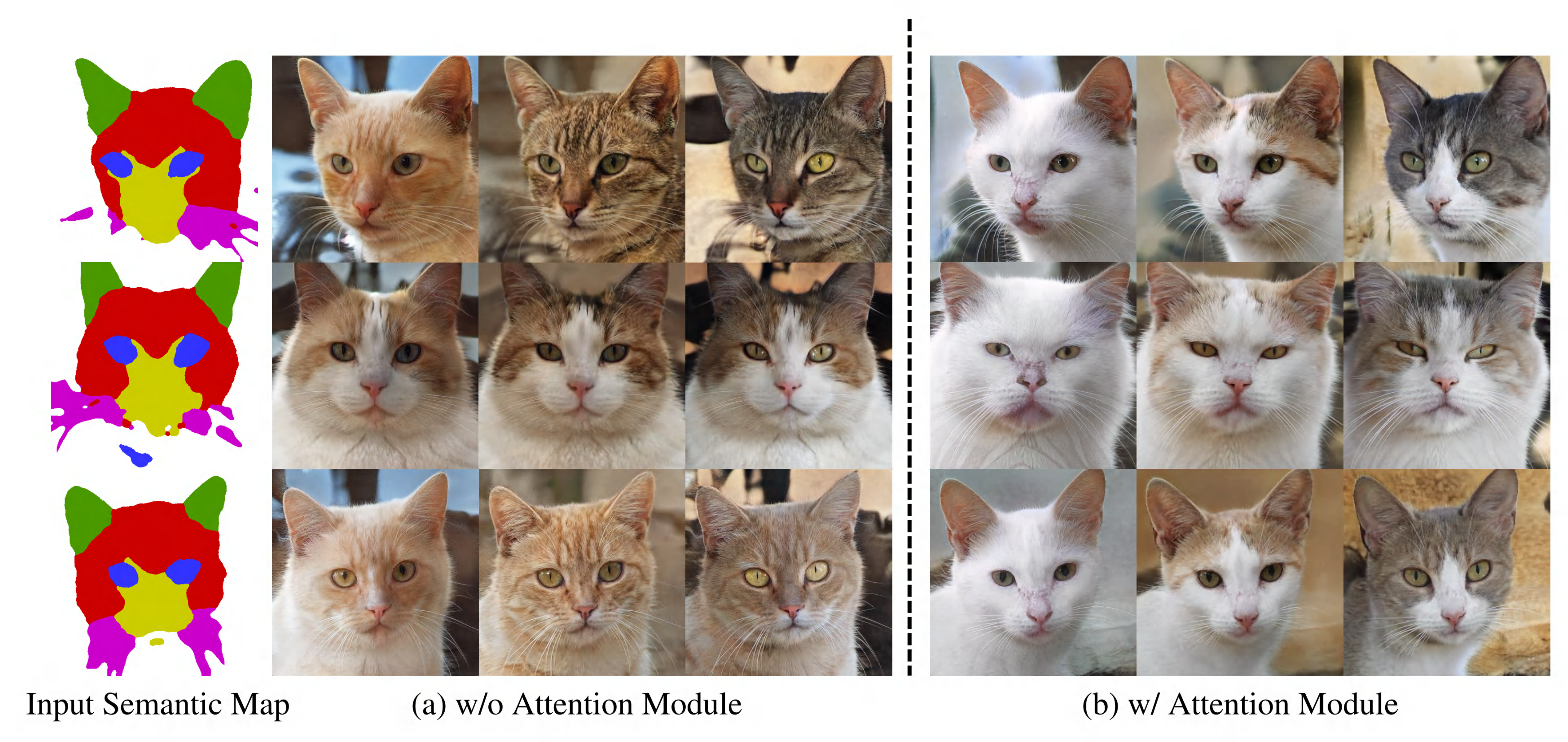}

 \vspace{-1em}
\caption{Experimental results of ablation study.
}

\label{fig:ablation}
\end{figure}

%% file: tab_seg2face.tex
\begin{table}[t!]
\caption{Quantitative Evaluation on CelebaMask dataset.}
\setlength{\tabcolsep}{1.5pt}
\setlength{\extrarowheight}{3pt}
\renewcommand{\arraystretch}{1}
\centering
\scriptsize
\begin{tabular}{lccc|cc|c|c}
\toprule
  \textsc{CELEBAMASK\cite{lee2020maskgan}}& \multicolumn{3}{c}{\textsc{Quality}} & \multicolumn{2}{c}{\textsc{Alignment}}& \multicolumn{1}{c}{\textsc{Identity}}&
  \multicolumn{1}{c}{\textsc{Style}}\\ %

 & FID $\downarrow$  & KID $\downarrow$ & Diversity $\uparrow$  & mIoU $\uparrow$ & acc $\uparrow$ & FVV $\downarrow$&   Gram $\downarrow$ \\ 
\midrule
\textsc{FENeRF \cite{sun2022fenerf}} & 56.73 & 0.047 & 0.26 & 0.47 & 0.81 & 0.55& 0.47\\
\textsc{SEAN \cite{zhu2020sean}} & 33.48 & 0.026 & 0.28 & 0.53 & 0.84 & N/A & 0.36\\

\textsc{SoFGAN \cite{chen2022sofgan}} & 28.64 & 0.019 & \textbf{0.33} & \textbf{0.55} & 0.87 & 0.63 &0.15  \\
\textsc{Pix2Pix3D \cite{deng20233d}} & 21.71 & 0.008 & 0.29 & 0.49 & 0.86 & 0.55 &0.40\\
\textsc{Ours } & \textbf{21.28} & \textbf{0.008} & 0.29 & 0.49 & \textbf{0.88}   &\textbf{0.53}&0.31 \\

\hline
\end{tabular}

\vspace{-1em}

\label{table:face_comparision}
\end{table}

%% file: tab_seg2cat.tex
\begin{table}[t!]
\caption{Quantitative Evaluation on AFHQ-Cat dataset.} 
\setlength{\tabcolsep}{4.3pt}
\setlength{\extrarowheight}{3pt}
\renewcommand{\arraystretch}{1}
\centering
\scriptsize
\begin{tabular}{lccc|cc|c}
\toprule
AFHQ-CAT\cite{choi2020starganv2}  & \multicolumn{3}{c}{\textsc{Quality}} & \multicolumn{2}{c}{\textsc{Alignment}}& \multicolumn{1}{c}{\textsc{Style}}\\ %

\textsc{} & FID $\downarrow$  & KID $\downarrow$ & Diversity $\uparrow$  & mIoU $\uparrow$ & acc $\uparrow$ $\downarrow$& Gram $\downarrow$ \\ 

\midrule

\textsc{SEAN \cite{zhu2020sean}} & 17.57 & 0.006 & 0.24 & 0.61 & 0.69 & 0.93 \\
\textsc{Pix2Pix3D \cite{deng20233d}} & 12.74 & 0.003 & 0.26 & 0.64 & 0.75 & 1.14 \\
\textsc{Ours } & \textbf{11.33} & \textbf{0.003} & \textbf{0.28} &\textbf{0.67}  & \textbf{0.77}   &\textbf{0.49} \\

\bottomrule
\end{tabular}


\label{table:cat_comparision}
\end{table}

%% file: fig_text_age.tex
\begin{figure}[t!]
    \centering
    \includegraphics[width=\linewidth]{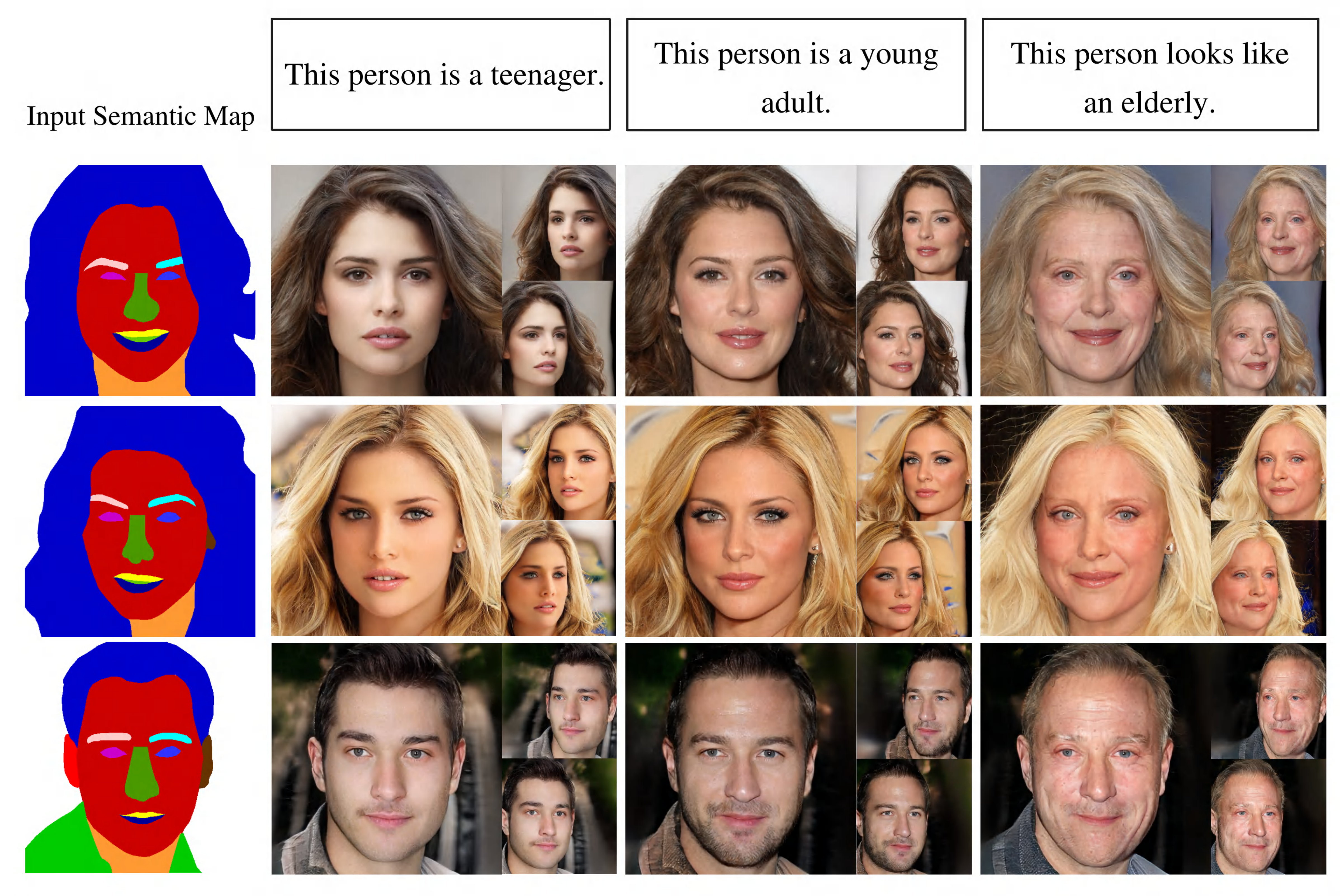}
    \includegraphics[width=\linewidth]{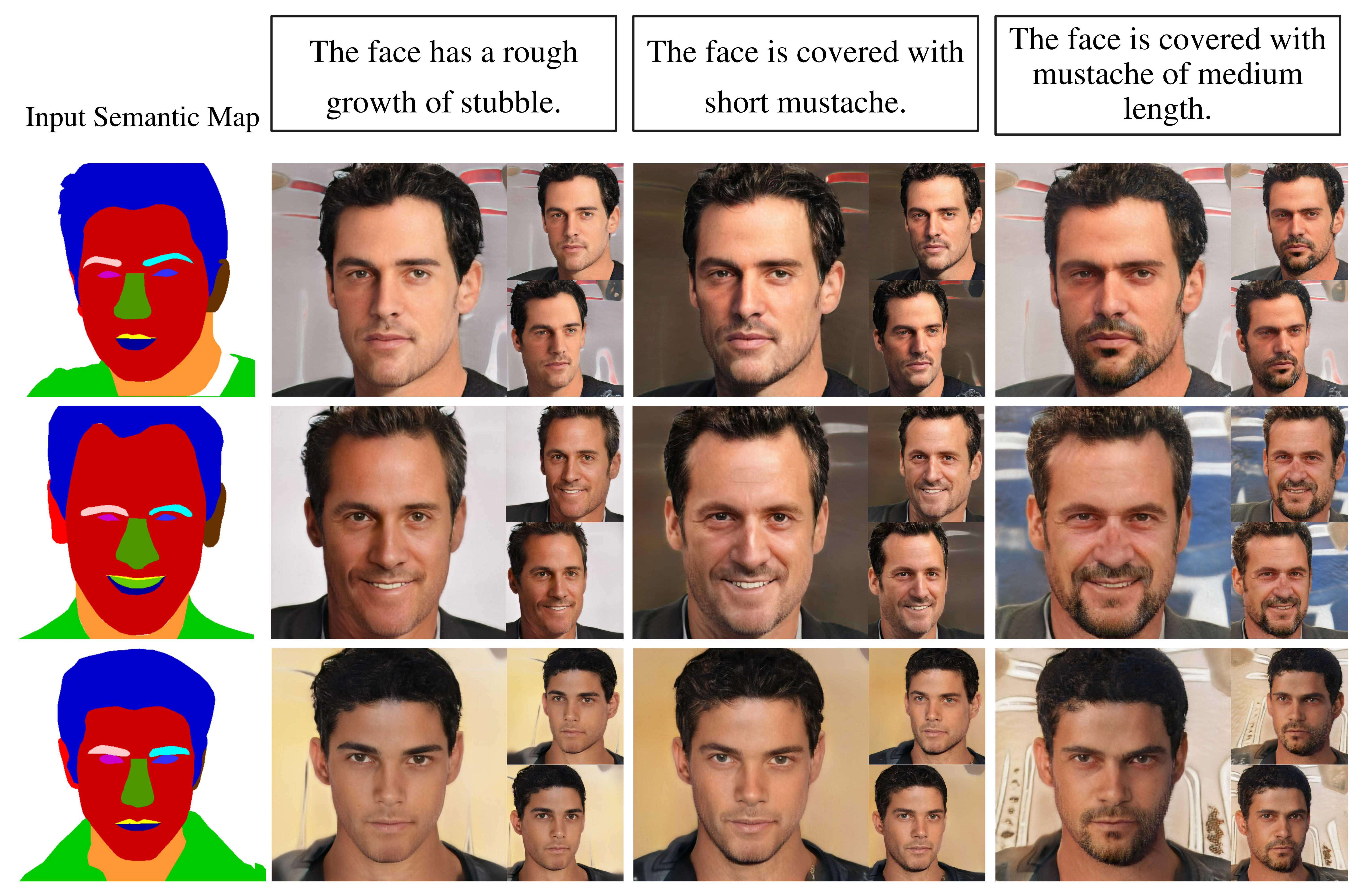}
    \vspace{-2em}
    \caption{Conditional generation and editing with text prompts. 
    }
    \label{fig:text_age}
    \vspace{-0.5em}
\end{figure}

%% file: fig_image_cat.tex
\begin{figure}[t!]
    \centering
    \includegraphics[width=\linewidth]{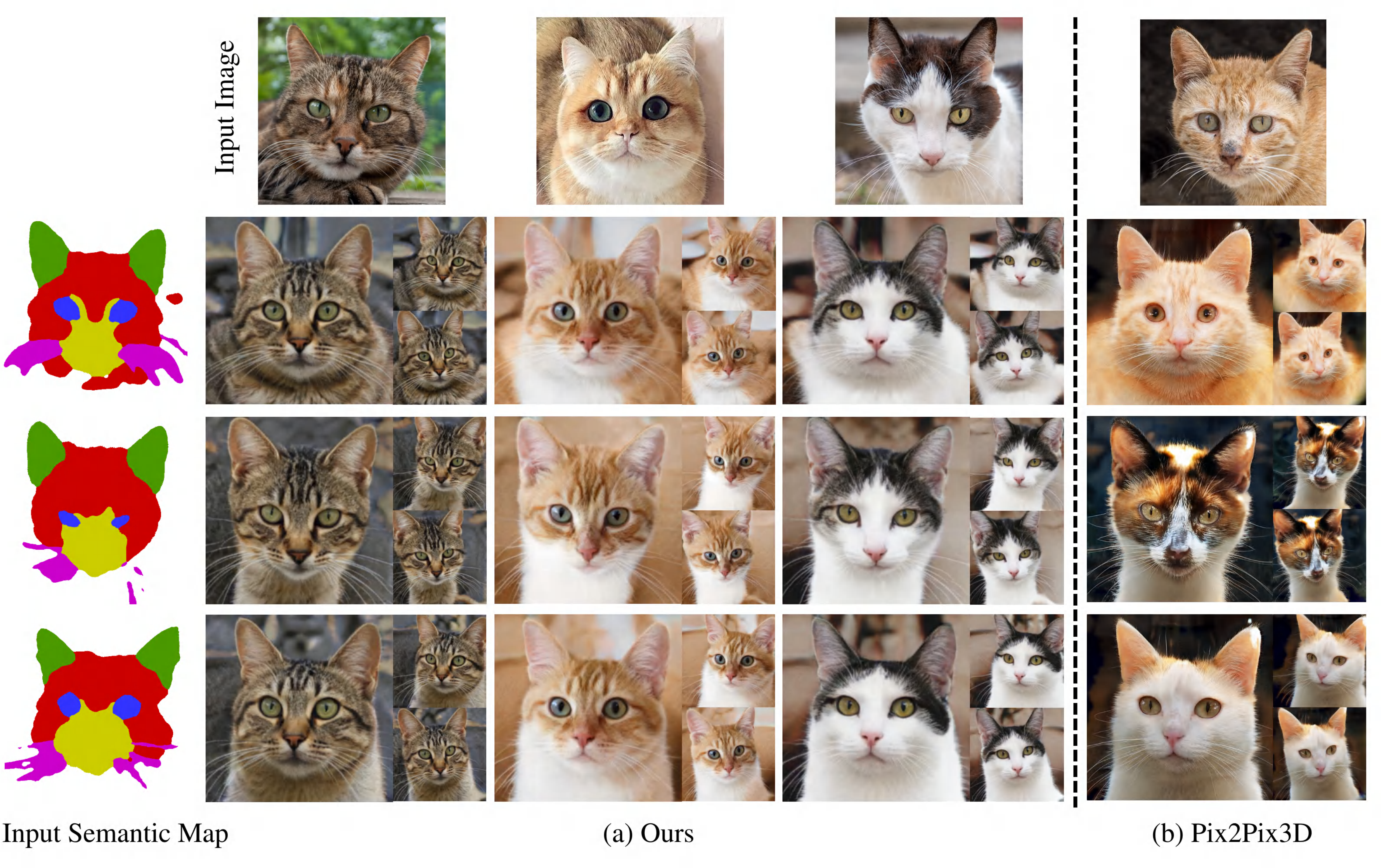}
    \vspace{-2em}
    \caption{\looseness=-1 Conditional generation results by utilizing a reference image as condition. 
    }
    \label{fig:image_cat}
    \vspace{-0.5em}
\end{figure}

%% file: fig_stylemix_compare.tex
\begin{figure*}[!t]
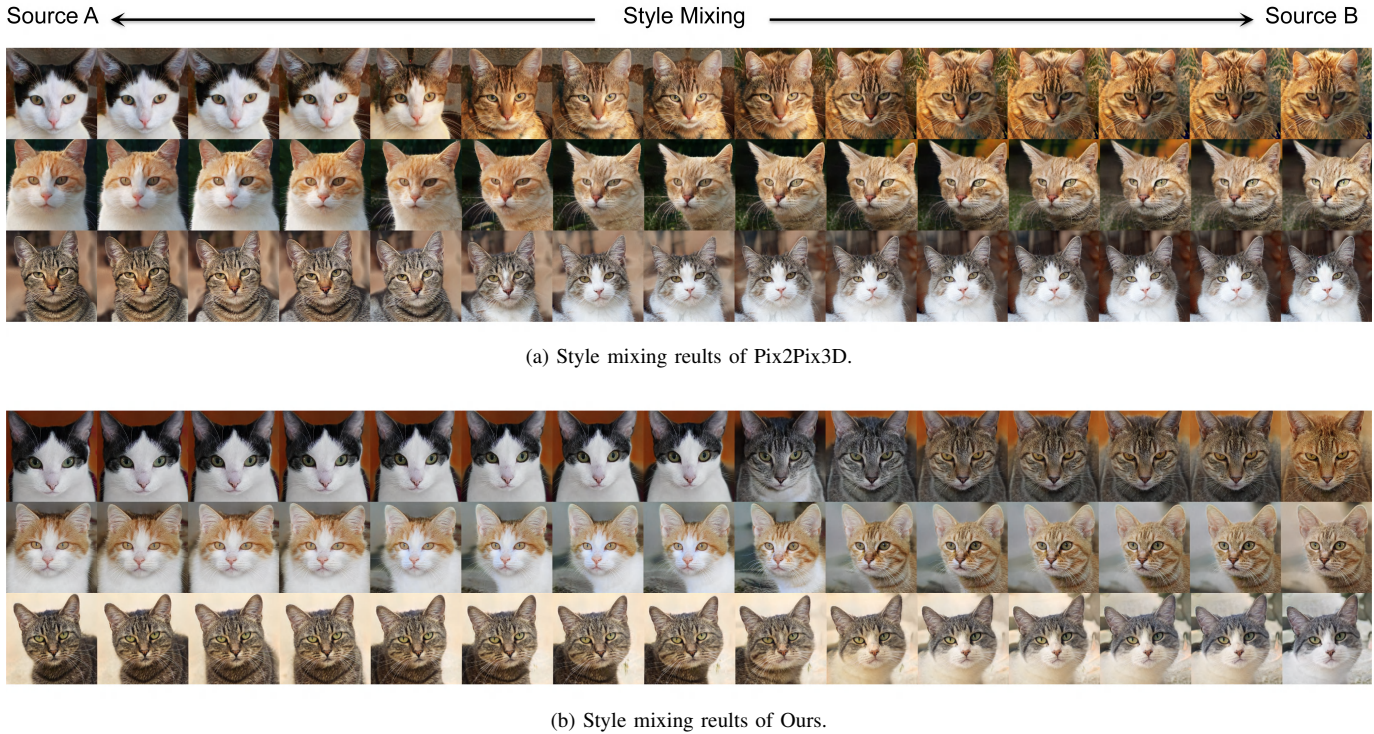

        \centering
        \vspace{0.5em}
        \subfloat[Style mixing reults of Pix2Pix3D.]{\includegraphics[width=\linewidth]{style_mixing_pix.pdf}}

        \subfloat[Style mixing reults of Ours.]{\includegraphics[width=\linewidth]{style_mixing_ours.pdf}}

    \caption{Latent space style mixing results of Pix2pix3D and the proposed method.}
    \label{fig:eg3d_style_mixing_compare}
\end{figure*}

%% file: fig_interp.tex
\begin{figure*}[t!]

\includegraphics[width=0.49\textwidth]{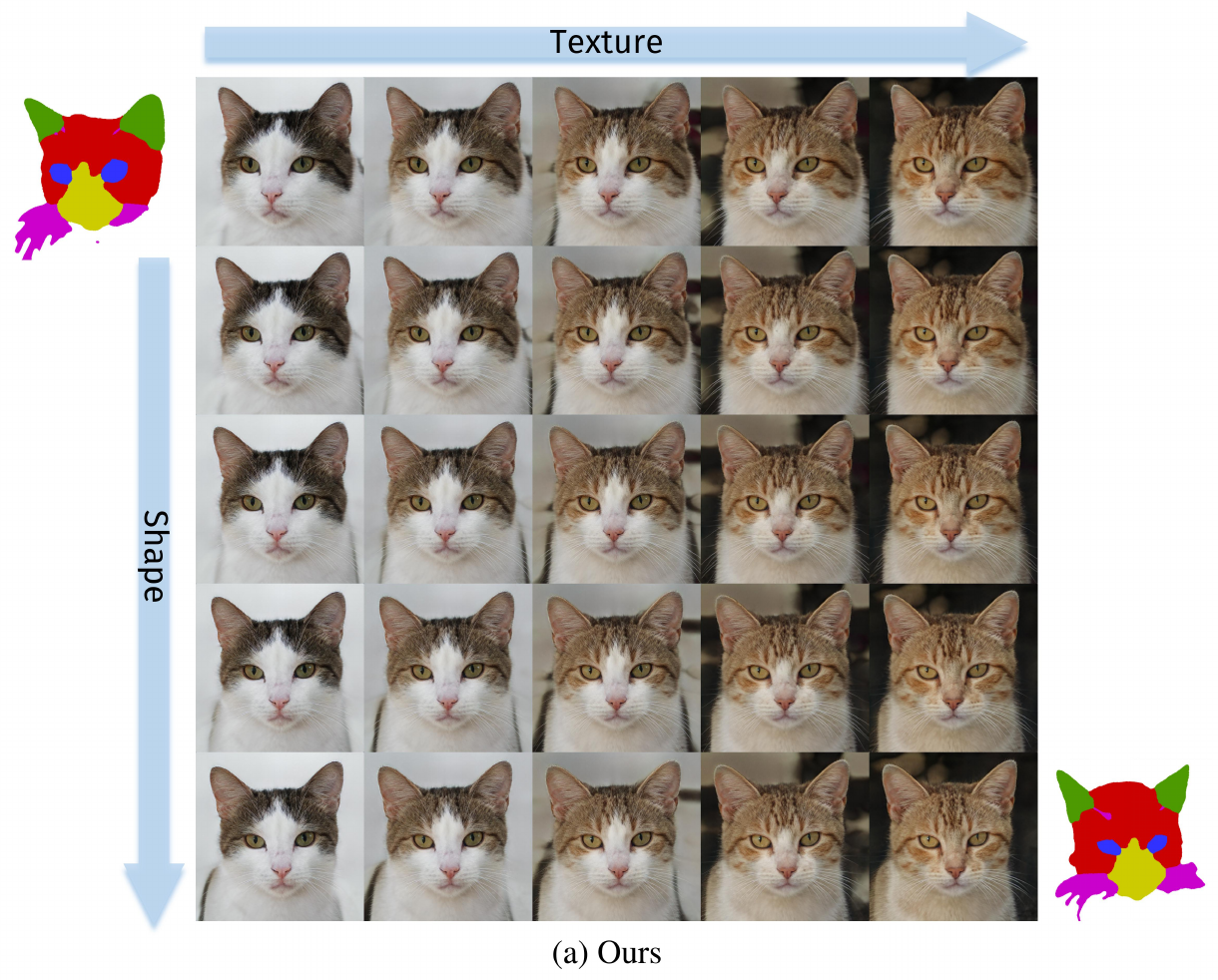}
\includegraphics[width=0.49\textwidth]{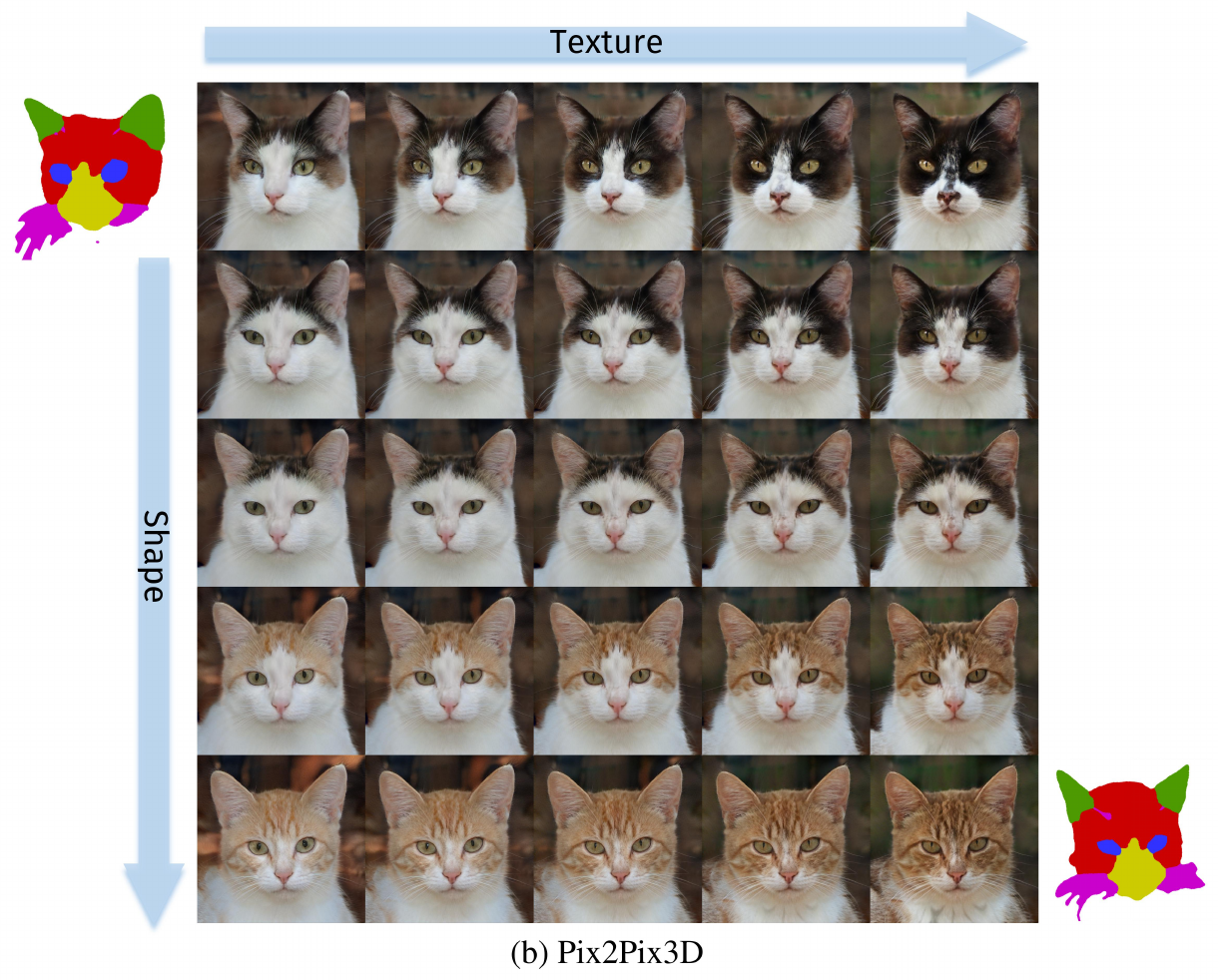}
 \vspace{-2em}
\caption{Results of latent space interpolation along shape feature and appearance feature. 
}
%

\label{fig:interp}
\end{figure*}

%% file: fig_3d.tex
\begin{figure}[!t]
        \includegraphics[width=\linewidth]{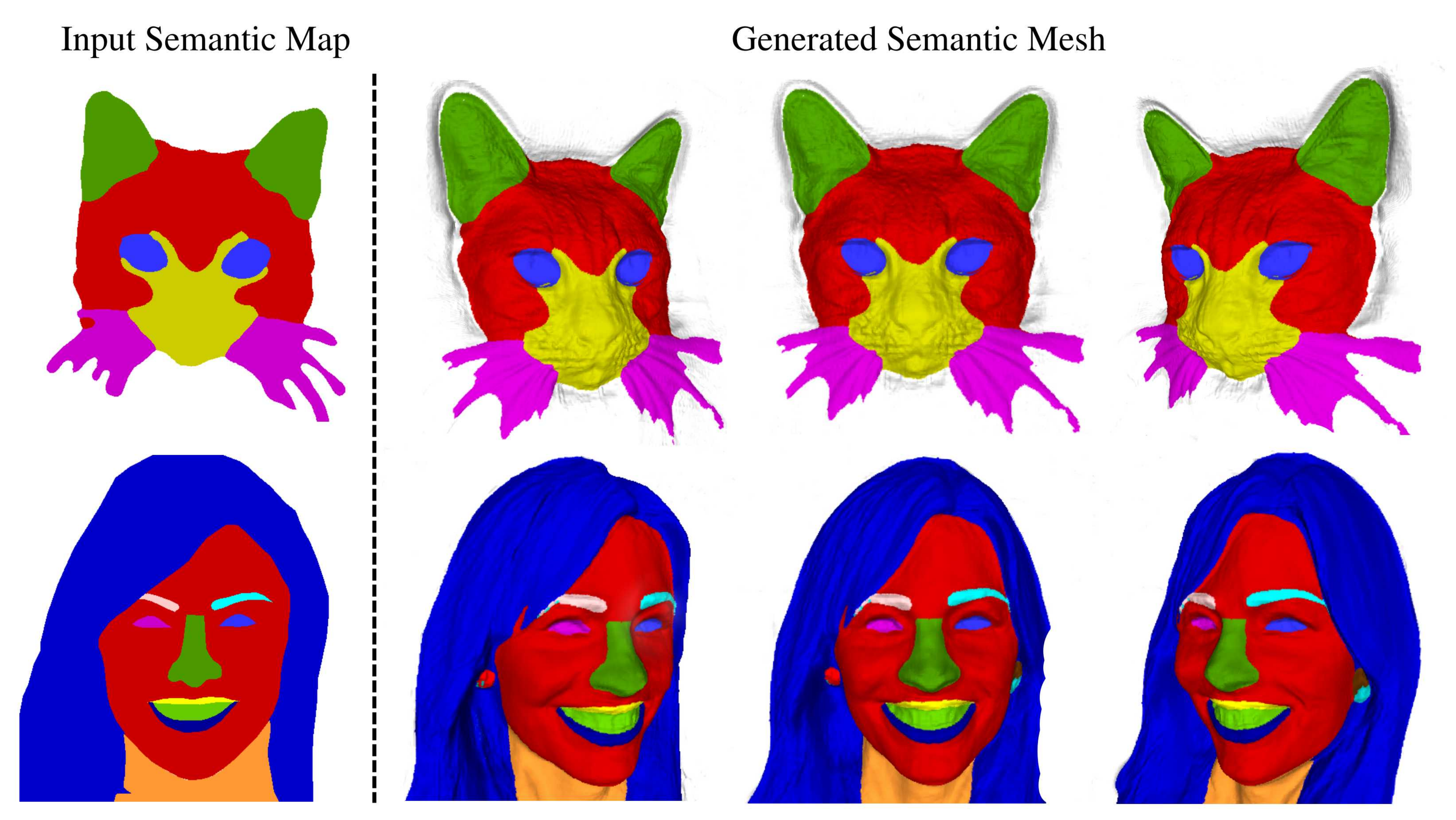}
    \vspace{-2em}
    \caption{3D semantic shape reconstruction results from as single semantic map. } 
    \vspace{-1.5em}
    \label{fig:3d}
\end{figure}

%% file: 5_conclusion.tex
This paper introduces an innovative approach for highly controllable 3D-aware image generation. It incorporates a novel disentanglement strategy that separates appearance features from shape features throughout the generation process, along with a unified framework for versatile image generation and editing tasks with multi-modal conditions. Extensive experiments verify its effectiveness in decoupling appearance from shape features, its flexible conditional controllability with multi-modal instructions, and its superiorities compared with state-of-the-art baselines in attribute editing. In future work, we will further investigate how to enhance 3D-aware generation with complex text instructions.



%% file: paper.bbl
\begin{thebibliography}{10}
\providecommand{\url}[1]{#1}
\csname url@samestyle\endcsname
\providecommand{\newblock}{\relax}
\providecommand{\bibinfo}[2]{#2}
\providecommand{\BIBentrySTDinterwordspacing}{\spaceskip=0pt\relax}
\providecommand{\BIBentryALTinterwordstretchfactor}{4}
\providecommand{\BIBentryALTinterwordspacing}{\spaceskip=\fontdimen2\font plus
\BIBentryALTinterwordstretchfactor\fontdimen3\font minus \fontdimen4\font\relax}
\providecommand{\BIBforeignlanguage}[2]{{%
\expandafter\ifx\csname l@#1\endcsname\relax
\typeout{** WARNING: IEEEtran.bst: No hyphenation pattern has been}%
\typeout{** loaded for the language `#1'. Using the pattern for}%
\typeout{** the default language instead.}%
\else
\language=\csname l@#1\endcsname
\fi
#2}}
\providecommand{\BIBdecl}{\relax}
\BIBdecl

\bibitem{shang2016understanding}
W.~Shang, K.~Sohn, D.~Almeida, and H.~Lee, ``Understanding and improving convolutional neural networks via concatenated rectified linear units,'' in \emph{international conference on machine learning}.\hskip 1em plus 0.5em minus 0.4em\relax PMLR, 2016, pp. 2217--2225.

\bibitem{goodfellow2020generative}
I.~Goodfellow, J.~Pouget-Abadie, M.~Mirza, B.~Xu, D.~Warde-Farley, S.~Ozair, A.~Courville, and Y.~Bengio, ``Generative adversarial networks,'' \emph{Communications of the ACM}, vol.~63, no.~11, pp. 139--144, 2020.

\bibitem{park2019semantic}
T.~Park, M.-Y. Liu, T.-C. Wang, and J.-Y. Zhu, ``Semantic image synthesis with spatially-adaptive normalization,'' in \emph{Proceedings of the IEEE/CVF conference on computer vision and pattern recognition}, 2019, pp. 2337--2346.

\bibitem{isola2017image}
P.~Isola, J.-Y. Zhu, T.~Zhou, and A.~A. Efros, ``Image-to-image translation with conditional adversarial networks,'' in \emph{Proceedings of the IEEE conference on computer vision and pattern recognition}, 2017, pp. 1125--1134.

\bibitem{chan2021efficient}
E.~R. Chan, C.~Z. Lin, M.~A. Chan, K.~Nagano, B.~Pan, S.~De~Mello, O.~Gallo, L.~Guibas, J.~Tremblay, S.~Khamis \emph{et~al.}, ``Efficient geometry-aware {3D} generative adversarial networks,'' \emph{arXiv preprint arXiv:2112.07945}, 2021.

\bibitem{Karras2020stylegan2}
T.~Karras, S.~Laine, M.~Aittala, J.~Hellsten, J.~Lehtinen, and T.~Aila, ``Analyzing and improving the image quality of {StyleGAN},'' in \emph{IEEE Conference on Computer Vision and Pattern Recognition (CVPR)}, 2020.

\bibitem{sun2022fenerf}
J.~Sun, X.~Wang, Y.~Zhang, X.~Li, Q.~Zhang, Y.~Liu, and J.~Wang, ``{FENeRF}: Face editing in neural radiance fields,'' in \emph{Proceedings of the IEEE/CVF Conference on Computer Vision and Pattern Recognition}, 2022, pp. 7672--7682.

\bibitem{zhang20223d}
J.~Zhang, E.~Sangineto, H.~Tang, A.~Siarohin, Z.~Zhong, N.~Sebe, and W.~Wang, ``{3D}-aware semantic-guided generative model for human synthesis,'' in \emph{European Conference on Computer Vision}.\hskip 1em plus 0.5em minus 0.4em\relax Springer, 2022, pp. 339--356.

\bibitem{rombach2022high}
R.~Rombach, A.~Blattmann, D.~Lorenz, P.~Esser, and B.~Ommer, ``High-resolution image synthesis with latent diffusion models,'' in \emph{Proceedings of the IEEE/CVF conference on computer vision and pattern recognition}, 2022, pp. 10\,684--10\,695.

\bibitem{poole2022dreamfusion}
B.~Poole, A.~Jain, J.~T. Barron, and B.~Mildenhall, ``{DreamFusion}: Text-to-{3D} using {2D} diffusion,'' \emph{arXiv preprint arXiv:2209.14988}, 2022.

\bibitem{wang2023score}
H.~Wang, X.~Du, J.~Li, R.~A. Yeh, and G.~Shakhnarovich, ``Score {Jacobian} chaining: Lifting pretrained {2D} diffusion models for {3D} generation,'' in \emph{Proceedings of the IEEE/CVF Conference on Computer Vision and Pattern Recognition}, 2023, pp. 12\,619--12\,629.

\bibitem{seo2023let}
J.~Seo, W.~Jang, M.-S. Kwak, J.~Ko, H.~Kim, J.~Kim, J.-H. Kim, J.~Lee, and S.~Kim, ``Let {2D} diffusion model know {3D}-consistency for robust text-to-{3D} generation,'' \emph{arXiv preprint arXiv:2303.07937}, 2023.

\bibitem{song2022diffusion}
K.~Song, L.~Han, B.~Liu, D.~Metaxas, and A.~Elgammal, ``Diffusion guided domain adaptation of image generators,'' \emph{arXiv preprint arXiv:2212.04473}, 2022.

\bibitem{deng20233d}
K.~Deng, G.~Yang, D.~Ramanan, and J.-Y. Zhu, ``{3D}-aware conditional image synthesis,'' in \emph{Proceedings of the IEEE/CVF Conference on Computer Vision and Pattern Recognition}, 2023, pp. 4434--4445.

\bibitem{graf}
K.~Schwarz, Y.~Liao, M.~Niemeyer, and A.~Geiger, ``{GRAF}: {G}enerative radiance fields for {3D}-aware image synthesis,'' in \emph{Advances in Neural Information Processing Systems (NeurIPS)}, 2020.

\bibitem{pigan}
E.~R. Chan, M.~Monteiro, P.~Kellnhofer, J.~Wu, and G.~Wetzstein, ``pi-gan: Periodic implicit generative adversarial networks for 3d-aware image synthesis,'' in \emph{Proceedings of the IEEE/CVF conference on computer vision and pattern recognition}, 2021, pp. 5799--5809.

\bibitem{Niemeyer2020GIRAFFE}
M.~Niemeyer and A.~Geiger, ``{GIRAFFE}: {R}epresenting scenes as compositional generative neural feature fields,'' in \emph{IEEE Conference on Computer Vision and Pattern Recognition (CVPR)}, 2021.

\bibitem{gu2021stylenerf}
J.~Gu, L.~Liu, P.~Wang, and C.~Theobalt, ``{StyleNeRF}: A style-based {3D}-aware generator for high-resolution image synthesis,'' \emph{arXiv preprint arXiv:2110.08985}, 2021.

\bibitem{zhang2022multi}
X.~Zhang, Z.~Zheng, D.~Gao, B.~Zhang, P.~Pan, and Y.~Yang, ``Multi-view consistent generative adversarial networks for {3D}-aware image synthesis,'' in \emph{Proceedings of the IEEE/CVF Conference on Computer Vision and Pattern Recognition}, 2022, pp. 18\,450--18\,459.

\bibitem{or2021stylesdf}
R.~Or-El, X.~Luo, M.~Shan, E.~Shechtman, J.~J. Park, and I.~Kemelmacher-Shlizerman, ``{StyleSDF}: High-resolution 3d-consistent image and geometry generation,'' \emph{arXiv e-prints}, pp. arXiv--2112, 2021.

\bibitem{shi2023learning}
Z.~Shi, Y.~Shen, Y.~Xu, S.~Peng, Y.~Liao, S.~Guo, Q.~Chen, and D.-Y. Yeung, ``Learning {3D}-aware image synthesis with unknown pose distribution,'' in \emph{Proceedings of the IEEE/CVF Conference on Computer Vision and Pattern Recognition}, 2023, pp. 13\,062--13\,071.

\bibitem{xu2018attngan}
T.~Xu, P.~Zhang, Q.~Huang, H.~Zhang, Z.~Gan, X.~Huang, and X.~He, ``Attn{GAN}: Fine-grained text to image generation with attentional generative adversarial networks,'' in \emph{Proceedings of the IEEE conference on computer vision and pattern recognition}, 2018, pp. 1316--1324.

\bibitem{zhang2017stackgan}
H.~Zhang, T.~Xu, H.~Li, S.~Zhang, X.~Wang, X.~Huang, and D.~N. Metaxas, ``Stack{GAN}: Text to photo-realistic image synthesis with stacked generative adversarial networks,'' in \emph{Proceedings of the IEEE international conference on computer vision}, 2017, pp. 5907--5915.

\bibitem{zhang2018stackgan++}
------, ``Stack{GAN}++: Realistic image synthesis with stacked generative adversarial networks,'' \emph{IEEE transactions on pattern analysis and machine intelligence}, vol.~41, no.~8, pp. 1947--1962, 2018.

\bibitem{zhu2017unpaired}
J.-Y. Zhu, T.~Park, P.~Isola, and A.~A. Efros, ``Unpaired image-to-image translation using cycle-consistent adversarial networks,'' in \emph{IEEE International Conference on Computer Vision (ICCV)}, 2017.

\bibitem{karras2019style}
T.~Karras, S.~Laine, and T.~Aila, ``A style-based generator architecture for generative adversarial networks,'' in \emph{IEEE Conference on Computer Vision and Pattern Recognition (CVPR)}, 2019.

\bibitem{Karras2021}
T.~Karras, M.~Aittala, S.~Laine, E.~H\"ark\"onen, J.~Hellsten, J.~Lehtinen, and T.~Aila, ``Alias-free generative adversarial networks,'' in \emph{Advances in Neural Information Processing Systems (NeurIPS)}, 2021.

\bibitem{shen2020interfacegan}
Y.~Shen, C.~Yang, X.~Tang, and B.~Zhou, ``{InterfaceGAN}: Interpreting the disentangled face representation learned by gans,'' \emph{IEEE transactions on pattern analysis and machine intelligence}, 2020.

\bibitem{abdal2021styleflow}
R.~Abdal, P.~Zhu, N.~J. Mitra, and P.~Wonka, ``{StyleFlow}: Attribute-conditioned exploration of {StyleGAN}-generated images using conditional continuous normalizing flows,'' \emph{ACM Transactions on Graphics (TOG)}, vol.~40, no.~3, pp. 1--21, 2021.

\bibitem{harkonen2020ganspace}
E.~H{\"a}rk{\"o}nen, A.~Hertzmann, J.~Lehtinen, and S.~Paris, ``{GANSpace}: Discovering interpretable {GAN} controls,'' \emph{arXiv preprint arXiv:2004.02546}, 2020.

\bibitem{shen2021closed}
Y.~Shen and B.~Zhou, ``Closed-form factorization of latent semantics in gans,'' in \emph{Proceedings of the IEEE/CVF Conference on Computer Vision and Pattern Recognition}, 2021, pp. 1532--1540.

\bibitem{patashnik2021styleclip}
O.~Patashnik, Z.~Wu, E.~Shechtman, D.~Cohen-Or, and D.~Lischinski, ``{StyleCLIP}: Text-driven manipulation of stylegan imagery,'' in \emph{Proceedings of the IEEE/CVF International Conference on Computer Vision}, 2021, pp. 2085--2094.

\bibitem{cheng2023efficient}
Y.~Cheng, F.~Yin, X.~Huang, X.~Yu, J.~Liu, S.~Feng, Y.~Yang, and Y.~Tang, ``Efficient text-guided {3D}-aware portrait generation with score distillation sampling on distribution,'' \emph{arXiv preprint arXiv:2306.02083}, 2023.

\bibitem{sun2022ide}
J.~Sun, X.~Wang, Y.~Shi, L.~Wang, J.~Wang, and Y.~Liu, ``{IDE-3D}: Interactive disentangled editing for high-resolution {3D}-aware portrait synthesis,'' \emph{arXiv preprint arXiv:2205.15517}, 2022.

\bibitem{blanz2023morphable}
V.~Blanz and T.~Vetter, ``A morphable model for the synthesis of {3D} faces,'' in \emph{Seminal Graphics Papers: Pushing the Boundaries, Volume 2}, 2023, pp. 157--164.

\bibitem{brunton2014review}
A.~Brunton, A.~Salazar, T.~Bolkart, and S.~Wuhrer, ``Review of statistical shape spaces for {3D} data with comparative analysis for human faces,'' \emph{Computer Vision and Image Understanding}, vol. 128, pp. 1--17, 2014.

\bibitem{bolkart2016robust}
T.~Bolkart and S.~Wuhrer, ``A robust multilinear model learning framework for {3D} faces,'' in \emph{Proceedings of the IEEE conference on computer vision and pattern recognition}, 2016, pp. 4911--4919.

\bibitem{booth20163d}
J.~Booth, A.~Roussos, S.~Zafeiriou, A.~Ponniah, and D.~Dunaway, ``A {3D} morphable model learnt from 10,000 faces,'' in \emph{Proceedings of the IEEE conference on computer vision and pattern recognition}, 2016, pp. 5543--5552.

\bibitem{cao2013facewarehouse}
C.~Cao, Y.~Weng, S.~Zhou, Y.~Tong, and K.~Zhou, ``Facewarehouse: A {3D} facial expression database for visual computing,'' \emph{IEEE Transactions on Visualization and Computer Graphics}, vol.~20, no.~3, pp. 413--425, 2013.

\bibitem{li2017learning}
T.~Li, T.~Bolkart, M.~J. Black, H.~Li, and J.~Romero, ``Learning a model of facial shape and expression from {4D} scans,'' \emph{ACM Trans. Graph.}, vol.~36, no.~6, pp. 194--1, 2017.

\bibitem{wang2017learning}
M.~Wang, Y.~Panagakis, P.~Snape, and S.~Zafeiriou, ``Learning the multilinear structure of visual data,'' in \emph{Proceedings of the IEEE conference on computer vision and pattern recognition}, 2017, pp. 4592--4600.

\bibitem{gecer2019ganfit}
B.~Gecer, S.~Ploumpis, I.~Kotsia, and S.~Zafeiriou, ``{GANFIT}: Generative adversarial network fitting for high fidelity {3D} face reconstruction,'' in \emph{Proceedings of the IEEE/CVF conference on computer vision and pattern recognition}, 2019, pp. 1155--1164.

\bibitem{lattas2020avatarme}
A.~Lattas, S.~Moschoglou, B.~Gecer, S.~Ploumpis, V.~Triantafyllou, A.~Ghosh, and S.~Zafeiriou, ``{AvatarMe}: Realistically renderable {3D} facial reconstruction ``in-the-wild''','' in \emph{Proceedings of the IEEE/CVF conference on computer vision and pattern recognition}, 2020, pp. 760--769.

\bibitem{gecer2020synthesizing}
B.~Gecer, A.~Lattas, S.~Ploumpis, J.~Deng, A.~Papaioannou, S.~Moschoglou, and S.~Zafeiriou, ``Synthesizing coupled {3D} face modalities by trunk-branch generative adversarial networks,'' in \emph{European conference on computer vision}.\hskip 1em plus 0.5em minus 0.4em\relax Springer, 2020, pp. 415--433.

\bibitem{gecer2018semi}
B.~Gecer, B.~Bhattarai, J.~Kittler, and T.-K. Kim, ``Semi-supervised adversarial learning to generate photorealistic face images of new identities from {3D} morphable model,'' in \emph{Proceedings of the European conference on computer vision (ECCV)}, 2018, pp. 217--234.

\bibitem{sela2017unrestricted}
M.~Sela, E.~Richardson, and R.~Kimmel, ``Unrestricted facial geometry reconstruction using image-to-image translation,'' in \emph{Proceedings of the IEEE international conference on computer vision}, 2017, pp. 1576--1585.

\bibitem{yu2018bisenet}
C.~Yu, J.~Wang, C.~Peng, C.~Gao, G.~Yu, and N.~Sang, ``{BiSeNet}: Bilateral segmentation network for real-time semantic segmentation,'' in \emph{Proceedings of the European conference on computer vision (ECCV)}, 2018, pp. 325--341.

\bibitem{amir2021deep}
S.~Amir, Y.~Gandelsman, S.~Bagon, and T.~Dekel, ``Deep vit features as dense visual descriptors,'' \emph{arXiv preprint arXiv:2112.05814}, vol.~2, no.~3, p.~4, 2021.

\bibitem{gatys2016image}
L.~A. Gatys, A.~S. Ecker, and M.~Bethge, ``Image style transfer using convolutional neural networks,'' in \emph{Proceedings of the IEEE conference on computer vision and pattern recognition}, 2016, pp. 2414--2423.

\bibitem{jiang2021talk}
Y.~Jiang, Z.~Huang, X.~Pan, C.~C. Loy, and Z.~Liu, ``Talk-to-edit: Fine-grained facial editing via dialog,'' in \emph{Proceedings of the IEEE/CVF International Conference on Computer Vision}, 2021, pp. 13\,799--13\,808.

\bibitem{lee2020maskgan}
C.-H. Lee, Z.~Liu, L.~Wu, and P.~Luo, ``Mask{GAN}: Towards diverse and interactive facial image manipulation,'' in \emph{Proceedings of the IEEE/CVF Conference on Computer Vision and Pattern Recognition}, 2020, pp. 5549--5558.

\bibitem{doosti2020hope}
B.~Doosti, S.~Naha, M.~Mirbagheri, and D.~J. Crandall, ``{HOPE-Net}: A graph-based model for hand-object pose estimation,'' in \emph{Proceedings of the IEEE/CVF conference on computer vision and pattern recognition}, 2020, pp. 6608--6617.

\bibitem{choi2020starganv2}
Y.~Choi, Y.~Uh, J.~Yoo, and J.-W. Ha, ``{StarGAN} v2: Diverse image synthesis for multiple domains,'' in \emph{Proceedings of the IEEE Conference on Computer Vision and Pattern Recognition}, 2020.

\bibitem{zhu2020sean}
P.~Zhu, R.~Abdal, Y.~Qin, and P.~Wonka, ``{SEAN}: Image synthesis with semantic region-adaptive normalization,'' in \emph{Proceedings of the IEEE/CVF Conference on Computer Vision and Pattern Recognition}, 2020, pp. 5104--5113.

\bibitem{chen2022sofgan}
A.~Chen, R.~Liu, L.~Xie, Z.~Chen, H.~Su, and J.~Yu, ``Sof{GAN}: A portrait image generator with dynamic styling,'' \emph{ACM Transactions on Graphics (TOG)}, vol.~41, no.~1, pp. 1--26, 2022.

\bibitem{heusel2017gans}
M.~Heusel, H.~Ramsauer, T.~Unterthiner, B.~Nessler, and S.~Hochreiter, ``{GANs} trained by a two time-scale update rule converge to a local {Nash} equilibrium,'' \emph{arXiv preprint arXiv:1706.08500}, 2017.

\bibitem{binkowski2018demystifying}
M.~Bińkowski, D.~J. Sutherland, M.~Arbel, and A.~Gretton, ``Demystifying {MMD} {GAN}s,'' in \emph{International Conference on Learning Representations (ICLR)}, 2018.

\bibitem{schroff2015facenet}
F.~Schroff, D.~Kalenichenko, and J.~Philbin, ``{FaceNet}: A unified embedding for face recognition and clustering,'' in \emph{Proceedings of the IEEE conference on computer vision and pattern recognition}, 2015, pp. 815--823.

\end{thebibliography}
